\documentclass[11pt]{article}

\usepackage{epsfig,epsf,fancybox}
\usepackage{amsmath}
\usepackage{mathrsfs}
\usepackage{amssymb}
\usepackage{graphicx}
\usepackage{color}
\usepackage{multirow}
\usepackage{paralist}
\usepackage{verbatim}
\usepackage{galois}
\usepackage{algorithm}
\usepackage{algorithmic}
\usepackage{boxedminipage}
\usepackage{booktabs}
\usepackage{accents}
\usepackage{stmaryrd}

\usepackage{subfig}

\usepackage{natbib}

\usepackage{url}% for url's in bib
\usepackage[colorlinks,linkcolor=magenta,citecolor=blue, pagebackref=true,backref=true]{hyperref}
\renewcommand*{\backrefalt}[4]{%
    \ifcase #1 \footnotesize{(Not cited.)}%
    \or        \footnotesize{(Cited on page~#2.)}%
    \else      \footnotesize{(Cited on pages~#2.)}%
    \fi}

\newtheorem{theorem}{Theorem}[section]

\newtheorem{lemma}[theorem]{Lemma}
\newtheorem{proposition}[theorem]{Proposition}

\newtheorem{definition}{Definition}[section]

\newtheorem{remark}[theorem]{Remark}
\newtheorem{assumption}[theorem]{Assumption}

\numberwithin{equation}{section}

\newcommand{\st}{\textnormal{s.t.}}

\newcommand{\Tr}{\textnormal{Tr}}
\newcommand{\conv}{\textnormal{conv}}

\newcommand{\diag}{\textnormal{diag}}

\newcommand{\QCal}{\mathcal{Q}}
\newcommand{\MCal}{\mathcal{M}}
\newcommand{\SCal}{\mathcal{S}}
\newcommand{\TCal}{\mathcal{T}}

\newcommand{\ICal}{\mathcal{I}}

\newcommand{\proj}{\textnormal{proj}}

\newcommand{\OT}{\textnormal{OT}}

\usepackage{mathrsfs}
\newcommand{\br}{\mathbb{R}}
\newcommand{\bs}{\mathbb{S}}

\newcommand{\ba}{\begin{array}}
\newcommand{\ea}{\end{array}}

\newcommand{\JCal}{\mathcal{J}}

\newcommand{\PScr}{\mathscr{P}}

\newcommand{\mydefn}{:=}

\begin{document}

%%%%%%% TITLE PAGE %%%%%%%%%%%%%%%%%%%%%%%%%%%%%%%%%%%%%%%%%%%%%%%%%%%

\begin{center}

{\bf{\LARGE{A Specialized Semismooth Newton Method for \\[.2cm] Kernel-Based Optimal Transport}}}

\vspace*{.2in}
{\large{ \begin{tabular}{c}
Tianyi Lin$^\ddagger$ \and Marco Cuturi$^{\triangleleft, \triangleright}$ \and Michael I. Jordan$^{\diamond, \dagger}$
\end{tabular}
}}

\vspace*{.2in}

\begin{tabular}{c}
Department of Electrical Engineering and Computer Sciences$^\diamond$ \\
Department of Statistics$^\dagger$ \\
University of California, Berkeley \\
Laboratory for Information and Decision Systems (LIDS), MIT$^\ddagger$ \\
CREST - ENSAE$^\triangleleft$, Apple$^\triangleright$
\end{tabular}

\vspace*{.2in}

\today

\vspace*{.2in}

\begin{abstract}
Kernel-based optimal transport (OT) estimators offer an alternative, functional estimation procedure to address OT problems from samples. Recent works suggest that these estimators are more statistically efficient than plug-in (linear programming-based) OT estimators when comparing probability measures in high-dimensions~\citep{Vacher-2021-Dimension}. Unfortunately, that statistical benefit comes at a very steep computational price: because their computation relies on the short-step interior-point method (SSIPM), which comes with a large iteration count in practice, these estimators quickly become intractable w.r.t. sample size $n$. To scale these estimators to larger $n$, we propose a nonsmooth fixed-point model for the kernel-based OT problem, and show that it can be efficiently solved via a specialized semismooth Newton (SSN) method: We show, exploring the problem's structure, that the per-iteration cost of performing one SSN step can be significantly reduced in practice. We prove that our SSN method achieves a global convergence rate of $O(1/\sqrt{k})$, and a local quadratic convergence rate under standard regularity conditions. We show substantial speedups over SSIPM on both synthetic and real datasets.
\end{abstract}

\end{center}

%!TEX root = ../paper.tex
\section{Introduction}\label{sec:intro}
Optimal transport (OT) theory~\citep{Santambrogio-2015-Optimal} provides a principled framework to compare probability distributions. OT has been used extensively in machine learning and related fields, notably for generative modeling~\citep{Montavon-2016-Wasserstein, Arjovsky-2017-Wasserstein, Genevay-2018-Learning, Salimans-2018-Improving, Tolstikhin-2018-Wasserstein}, classification and clustering~\citep{Frogner-2015-Learning, Srivastava-2015-WASP, Ho-2017-Multilevel}, or domain adaptation~\citep{Courty-2016-Optimal, Courty-2017-Joint, Redko-2019-Optimal}, see~\citet{Peyre-2019-Computational}. OT is also impactful in applied areas such as neuroimaging~\citep{Janati-2020-Multi} and cell trajectory prediction~\citep{Schiebinger-2019-Optimal, Yang-2020-Predicting}.

\paragraph{Curse of Dimensionality.} In most applications, the OT problem is seeded with the squared Euclidean distance as the ground cost, and instantiated with $n$ samples. In that regime, OT estimation is known to suffer from the curse of dimensionality~\citep{Dudley-1969-Speed, Fournier-2015-Rate, Weed-2019-Sharp}: The standard plug-in estimator for the OT objective, which runs a linear program on those samples, converges to its population value at a rate of $O(n^{-2/d})$~\citep{Chizat-2020-Faster}, hindering the adoption of OT in machine learning. Practitioners are aware of such limitations and use alternative computational schemes that improve computational complexity while carrying out statistical regularization. 

\paragraph{Regularization.} Quite a few works propose to regularize the OT problem: using entropy~\citep{Cuturi-2013-Sinkhorn,Genevay-2019-Sample, Mena-2019-Statistical}, low-dimensional projections~\citep{Rabin-2011-Wasserstein, Bonneel-2015-Sliced, Paty-2019-Subspace, Kolouri-2019-Generalized, Nadjahi-2020-Statistical, Lin-2020-Projection, Lin-2021-Projection, Weed-2022-Estimation}, bootstrap~\citet{Sommerfeld-2018-Inference,Fatras-2020-Learning} or neural networks~\citep{Amos-2017-Input,Makkuva-2020-Optimal, Korotin-2021-Wasserstein}. The sample complexity of entropic OT is bounded by $O(\varepsilon^{-d/2}n^{-1/2})$ for a regularization strength $\varepsilon > 0$, while that of projected OT is bounded by $O(n^{-1/k})$ for projection dimension $k \leq d$. Although these bounds may seem dimension-free w.r.t. $n$, they deteriorate when $\eta$ is small or $k$ is large, losing relevance to the original OT problem~\citep{Chizat-2020-Faster}. Minibatch approaches are mostly used as fitting loss, while neural approaches are used with no guarantees.

\paragraph{Leveraging Smoothness.} Alternative approaches build on strong smoothness assumptions on potentials or maps, such as \textit{wavelet}-based estimators~\citep{Weed-2019-Estimation, Hutter-2021-Minimax, Deb-2021-Rates, Manole-2021-Plugin}, which are minimax optimal but algorithmically intractable. These approaches stand in contrast to, e.g., entropic map estimators~\citep{Pooladian-2021-Entropic} which are cheap but still suffer from the curse of dimensionality. Recently,~\citet{Vacher-2021-Dimension} closed this statistical-computational gap by designing an estimator that relies on kernel sums-of-squares, showing that it can be computed using a short-step interior-point method (SSIPM), with polynomial-time complexity guarantee. Unfortunately, the SSIPM is known to be ineffective~\citet{Potra-2000-Interior}, requiring a large number of iterations as sample size grows. This issue was specifically pointed out in~\citep[p.11-12]{Vacher-2021-Dimension}, and we do observe it experimentally (see Fig.~\ref{fig:time}). 

\paragraph{Scaling up Kernel-based OT.} While~\citeauthor{Vacher-2021-Dimension}'s method holds several promises on the statistical front, it does lack an efficient implementation. Such an implementation is needed if one wants to show that these theoretical benefits do translate into practical advantages. \citet{Muzellec-2021-Near} proposed to improve this computational outlook with an additional relaxation. Their mollified problem can be solved with simple gradient-based methods, but presents, however, a significant departure from the original kernel-based estimator and its guarantees. We follow in their footsteps but focus directly on improving the computational efficiency of~\citeauthor{Vacher-2021-Dimension}'s estimator. We address~\citeauthor{Vacher-2021-Dimension}'s original problem using the semismooth Newton (SSN) method~\citep{Mifflin-1977-Semismooth, Qi-1993-Nonsmooth, Qi-1999-Survey, Ulbrich-2011-Semismooth}. Our contribution is therefore purely \textit{computational}: Since our approach targets the same optimization problem, our estimators inherit the statistical guarantees proved in~\citep{Vacher-2021-Dimension}. Note that SSN methods were recently used in an OT context by~\citet{Liu-2022-Multiscale}, but in the unrelated setting of solving the multiscale min-cost-flow problem on grids.

\paragraph{Contributions.} We propose a nonsmooth equation model for kernel-based OT problems. We use it to devise a specialized SSN method to compute kernel-based OT estimators, and prove a global rate of $O(1/\sqrt{k})$ (Theorem~\ref{Thm:convergence-global}) and a local quadratic rate under standard regularity conditions (Theorem~\ref{Thm:convergence-local}). We show how to significantly reduce the per-iteration cost of our algorithm by exploiting structure. Finally, we validate experimentally that SSN is substantially faster than SSIPM on both synthetic and real data, and use our estimators to produce OT (Monge) map estimators, benchmarked on single-cell data.

\paragraph{Organization.} The remainder is organized as follows. In Section~\ref{sec:prelim}, we present the nonsmooth equation model for computing the kernel-based OT estimators and define the optimality notion based on the residual map. In Section~\ref{sec:alg}, we propose and analyze the specialized SSN algorithm for computing the kernel-based OT estimators and prove that our algorithm achieves the global and local convergence rate guarantees. In Section~\ref{sec:exp}, we conduct the experiments on both synthetic and real datasets, demonstrating that our algorithm can effectively compute the kernel-based OT estimators and is more efficient than short-step interior-point methods. In Section~\ref{sec:conclu}, we conclude this paper. In the supplementary material, we provide additional experimental results, and missing proofs for key results. 

\section{Further Related Works}
Semismooth Newton (SSN) methods~\citep{Ulbrich-2011-Semismooth} are a class of powerful and versatile algorithms for solving constrained optimization problems with PDEs, and variational inequalities (VIs). The notion of semi-smoothness was introduced by~\citet{Mifflin-1977-Semismooth} for real-valued functions and then extended to vector-valued mappings by~\citet{Qi-1993-Nonsmooth}. A pioneering work on the SSN method was due to~\citet{Solodov-1999-Globally}, in which the authors proposed a globally convergent Newton method by exploiting the structure of monotonicity and established a local superlinear convergence rate under the conditions that the generalized Jacobian is semismooth and nonsingular at the global optimal solution. The convergence rate guarantee was later extended in~\citet{Zhou-2005-Superlinear} to the setting where the generalized Jacobian is not nonsingular. 

The SSN methods have received significant amount of attention due to its wide success in solving several structured convex problems to a high accuracy. In particular, such approach has been successfully applied to solving large-scale SDPs~\citep{Zhao-2010-Newton, Yang-2015-SDPNAL}, LASSO~\citep{Li-2018-Highly}, nearest correlation matrix estimation~\citep{Qi-2011-Augmented}, clustering~\citep{Wang-2010-Solving}, sparse inverse covariance selection~\citep{Yang-2013-Proximal} and composite convex minimization~\citep{Xiao-2018-Regularized}. The closest works to ours is~\citet{Liu-2022-Multiscale}, who developed a fast SSN method to compute the plug-in OT estimator by exploring the sparsity and multiscale structure of its linear programming (LP) formulation. All of their experiments are run on 2D image grids. In contrast, our methods uses SSN to target a regularized, dual RKHS (functional) formulation, useful in higher dimensions. To our knowledge, this paper is the first to apply the SSN method to computing the kernel-based OT estimator and prove the convergence rate guarantees. 

%!TEX root = main.tex
\section{Background: Kernel-Based OT}\label{sec:prelim}
We formally define the OT problem and review the kernel-based OT estimator proposed by~\citet{Vacher-2021-Dimension}. Let $X$ and $Y$ be two bounded domains in $\br^d$ and let $\PScr(X)$ and $\PScr(Y)$ be the set of Borel probability measures in $X$ and $Y$. Suppose that $\mu \in \PScr(X)$, $\nu \in \PScr(Y)$ and $\Pi(\mu, \nu)$ is the set of couplings between $\mu$ and $\nu$, the primal OT problem is:
\begin{equation*}
\OT(\mu, \nu) \mydefn \tfrac{1}{2}\left(\inf_{\pi \in \Pi(\mu, \nu)} \int_{X \times Y} \|x-y\|^2 \ d\pi(x, y)\right)\,,
\end{equation*}
while its dual formulation is stated as follows, 
\begin{equation*}
\begin{array}{cl}
\sup\limits_{u, v \in C(\br^d)} & \int_X u(x) d\mu(x) + \int_Y v(y) d\nu(y), \\ 
\st & \tfrac{1}{2}\|x - y\|^2 \geq u(x) + v(y), \forall (x, y) \in X \times Y, 
\end{array}
\end{equation*}
where $C(\br^d)$ is the set of continuous functions on $\br^d$. Note that the supremum can be attained and the corresponding optimal dual functions $u_\star$ and $v_\star$ are referred to as the Kantorovich potentials~\citep{Santambrogio-2015-Optimal}. This problem has a continuous constraint set, since $\frac{1}{2}\|x - y\|^2 \geq u(x) + v(y)$ must be satisfied on $X \times Y$. A natural approach is to take $n$ points $\{(\tilde{x}_1, \tilde{y}_1), \ldots, (\tilde{x}_n, \tilde{y}_n)\} \subseteq X \times Y$ and consider the constraints $\frac{1}{2}\|\tilde{x}_i - \tilde{y}_i\|^2 \geq u(\tilde{x}_i) + v(\tilde{y}_i)$ for all $1 \leq i \leq n$. However, it can not leverage the smoothness of potentials~\citep{Aubin-2020-Hard}, yielding an error of $\Omega(n^{-1/d})$.~\citeauthor{Vacher-2021-Dimension} overcome this difficulty by replacing the inequality constraints with equivalent equality constraints, and considering these constraints over $n$ points. Following their work, we use the following assumptions on the support sets $X, Y$ and the densities of $\mu$ and $\nu$. 
\begin{assumption}\label{Assumption:smoothness}
Let $d \geq 1$ be the dimension and let $m > 2d+2$ be the order of smoothness. Then, we assume that (i) the support sets $X, Y$ are convex, bounded, and open with Lipschitz boundaries; (ii) the densities of $\mu, \nu$ are finite, bounded away from zero and $m$-times differentiable. 
\end{assumption}
Assumption~\ref{Assumption:smoothness} guarantees that the potentials $u_\star$ and $v_\star$ have a similar order of differentiability~\citep{De-2014-Monge}, leading to an effective way to represent $u$ and $v$ via a \textit{reproducing Kernel Hilbert space} (RKHS). In particular, we define $H^s(Z):=\{f \in L^2(Z) \mid \|f\|_{H^s(Z)} := \sum_{|\alpha| \leq s} \|D^\alpha f\|_{L^2(Z)} < +\infty\}$ and remark that $H^s(Z) \subseteq C^k(Z)$ for any $s > \frac{d}{2}+k$, where $k \geq 0$ is integer-valued. This guarantees that $H^{m+1}(X)$, $H^{m+1}(Y)$ and $H^m(X \times Y)$ are RKHS under Assumption~\ref{Assumption:smoothness}~\citep{Paulsen-2016-Introduction} and they are associated with three bounded continuous feature maps $\phi_X: X \mapsto H^{m+1}(X)$, $\phi_Y: Y \mapsto H^{m+1}(Y)$ and $\phi_{XY}: X \times Y \mapsto H^m(X \times Y)$. For simplicity, we let $H_X = H^{m+1}(X)$, $H_Y = H^{m+1}(Y)$ and $H_{XY} = H^m(X \times Y)$.~\citet[Corollary~7]{Vacher-2021-Dimension} shows that (i) $u_\star \in H_X$ and $v_\star \in H_Y$ with 
\begin{equation*}
\int_X u(x) d\mu(x) = \langle u, w_\mu\rangle_{H_X}, \ \int_X v(y) d\nu(y) = \langle v, w_\nu\rangle_{H_Y}, 
\end{equation*}
where $w_\mu = \int_X \phi_X(x)d\mu(x)$ and $w_\nu = \int_Y \phi_Y(y)d\nu(y)$ are \textit{kernel mean embeddings}; (ii) $A_\star \in \bs^+(H_{XY})$\footnote{We refer to $\bs^+(H_{XY})$ as the set of linear, positive and self-adjoint operators on $H_{XY}$.} exists and satisfies the equality constraint as follows: 
\begin{equation*}
\tfrac{1}{2}\|x - y\|^2 - u_\star(x) - v_\star(y) = \left\langle \phi_{XY}(x, y), A_\star\phi_{XY}(x, y)\right\rangle_{H_{XY}}. 
\end{equation*}
Putting these pieces yields a representation theorem for estimating the OT distance. Indeed, under Assumption~\ref{Assumption:smoothness}, the dual OT problem is equivalent to the RKHS-based problem given by 
\begin{equation}\label{prob:OT-population}
\begin{array}{lcl}
& \max\limits_{u, v, A} & \langle u, w_\mu\rangle_{H_X} + \langle v, w_\nu\rangle_{H_Y}, \\
& \st & \tfrac{1}{2}\|x - y\|^2 - u(x) - v(y) = \langle \phi_{XY}(x, y), A\phi_{XY}(x, y)\rangle_{H_{XY}}. 
\end{array}
\end{equation}
The above equation offers two advantages: (i) The equality constraint can be well approximated under Assumption~\ref{Assumption:smoothness}; (ii) RKHSs allow the kernel trick: computing parameters are expressed in terms of \textit{kernel functions} that correspond to
\begin{equation*}
\begin{array}{lcl}
k_X(x, x') & = & \langle \phi_X(x), \phi_X(x')\rangle_{H_X}, \\ 
k_Y(y, y') & = & \langle \phi_Y(y), \phi_Y(y')\rangle_{H_Y}, \\
k_{XY}((x, y), (x', y')) & = & \langle \phi_{XY}(x, y), \phi_{XY}(x', y')\rangle_{H_{XY}}, 
\end{array}
\end{equation*}
where the kernel functions are explicit and can be computed in $O(d)$ given the samples. The final step is to approximate Eq.~\eqref{prob:OT-population} using the data $x_1, \ldots, x_{n_\textnormal{sample}} \sim \mu$ and $y_1, \ldots, y_{n_\textnormal{sample}} \sim \nu$, and the filling points $\{(\tilde{x}_1, \tilde{y}_1), \ldots, (\tilde{x}_n, \tilde{y}_n)\} \subseteq X \times Y$. Indeed, we define $\hat{\mu} = \tfrac{1}{n_\textnormal{sample}}\sum_{i=1}^{n_\textnormal{sample}} \delta_{x_i}$ and $\hat{\nu} = \tfrac{1}{n_\textnormal{sample}}\sum_{i=1}^{n_\textnormal{sample}} \delta_{y_i}$, and use $\langle u, w_{\hat{\mu}}\rangle_{H_X} + \langle v, w_{\hat{\nu}}\rangle_{H_Y}$ instead of $\langle u, w_\mu\rangle_{H_X} + \langle v, w_\nu\rangle_{H_Y}$ where $w_{\hat{\mu}} = \frac{1}{n_\textnormal{sample}}\sum_{i=1}^{n_\textnormal{sample}} \phi_X(x_i)$ and $w_{\hat{\nu}} = \frac{1}{n_\textnormal{sample}}\sum_{i=1}^{n_\textnormal{sample}} \phi_Y(y_i)$. We also impose \textit{the penalization terms} for $u$, $v$, and $A$ to alleviate the error induced by sampling the corresponding equality constraints. Then, the resulting problem with regularization parameters $\lambda_1, \lambda_2 > 0$ is summarized as follows: 
\begin{equation}\label{prob:W2-empirical}
\begin{array}{lcl}
& \max\limits_{u, v, A} & \langle u, w_{\hat{\mu}}\rangle_{H_X} + \langle v, w_{\hat{\nu}}\rangle_{H_Y} - \lambda_1\Tr(A) - \lambda_2(\|u\|_{H_X}^2 + \|v\|_{H_Y}^2), \\
& \st & \tfrac{1}{2}\|\tilde{x}_i - \tilde{y}_i\|^2 - u(\tilde{x}_i) - v(\tilde{y}_i) = \langle \phi_{XY}(\tilde{x}_i, \tilde{y}_i), A\phi_{XY}(\tilde{x}_i, \tilde{y}_i)\rangle_{H_{XY}}. 
\end{array}
\end{equation}
Focusing on the case of $n_\textnormal{sample} = \Theta(n)$, we let $\hat{u}_\star$ and $\hat{v}_\star$ be the unique maximizers of Eq.~\eqref{prob:W2-empirical}. Then, the estimator for $\OT(\mu, \nu)$ we consider corresponds to
\begin{equation}\label{def:estimator}
\widehat{\OT}^n = \langle \hat{u}_\star, w_{\hat{\mu}}\rangle_{H_X} + \langle \hat{v}_\star, w_{\hat{\nu}}\rangle_{H_Y}. 
\end{equation}
\begin{remark}
It follows from~\citet[Corollary~3]{Vacher-2021-Dimension} that the norm of empirical potentials can be controlled using $\lambda_1 = \tilde{\Theta}(n^{-1/2})$ and $\lambda_2 = \tilde{\Theta}(n^{-1/2})$ in high probability and the statistical rate is $\tilde{O}(n^{-1/2})$. Compared with plug-in OT estimators, the kernel-based OT estimators are better when sample size $n$ is small (estimator is still tractable) and dimension $d$ is large (statistical rates are $O(n^{-2/d})$ and $\tilde{O}(n^{-1/2})$ for plug-in and kernel-based estimators, respectively).
\end{remark}
\begin{remark}
The entropic OT estimators achieve the rate of $\tilde{O}(n^{-1/2})$ for fixed $\varepsilon$~\citep{Genevay-2019-Sample}. Such a rate blows up exponentially fast to infinity as $\varepsilon \rightarrow 0$ if one wants to approximate non-regularized OT. Hence, entropic OT estimators are only statistically efficient for fixed, and fairly large, values of $\varepsilon$. In contrast, kernel-based OT estimators do not suffer from such a blow-up. While the constants depend exponentially in $d$, they are fixed, and the rate of $\tilde{O}(n^{-1/2})$ is valid for approximating non-regularized OT.
\end{remark}
Eq.~\eqref{prob:W2-empirical} is an infinite-dimensional problem and is thus difficult to solve. Thanks to~\citet[Theorem~15]{Vacher-2021-Dimension}, we have that the dual problem of Eq.~\eqref{prob:W2-empirical} can be presented in a finite-dimensional space and strong duality holds true. Indeed, we define $Q \in \br^{n \times n}$ with $Q_{ij} = k_X(\tilde{x}_i, \tilde{x}_j) + k_Y(\tilde{y}_i, \tilde{y}_j)$, and $z \in \br^n$ with $z_i = w_{\hat{\mu}}(\tilde{x}_i) + w_{\hat{\nu}}(\tilde{y}_i) - \lambda_2\|\tilde{x}_i - \tilde{y}_i\|^2$, and $q^2 = \|w_{\hat{\mu}}\|_{H_X}^2 + \|w_{\hat{\nu}}\|_{H_Y}$, where we have
\begin{equation*}
\begin{array}{lcl}
w_{\hat{\mu}}(\tilde{x}_i) & = & \tfrac{1}{n_\textnormal{sample}}\sum_{j=1}^{n_\textnormal{sample}} k_X(x_j, \tilde{x}_i), \\ 
w_{\hat{\nu}}(\tilde{y}_i) & = & \tfrac{1}{n_\textnormal{sample}}\sum_{j=1}^{n_\textnormal{sample}} k_Y(y_j, \tilde{y}_i), \\
\|w_{\hat{\mu}}\|_{H_X}^2 & = & \tfrac{1}{n_\textnormal{sample}^2}\sum_{1 \leq i, j \leq n_\textnormal{sample}} k_X(x_i, x_j), \\
\|w_{\hat{\nu}}\|_{H_Y}^2 & = & \tfrac{1}{n_\textnormal{sample}^2}\sum_{1 \leq i, j \leq n_\textnormal{sample}} k_Y(y_i, y_j).
\end{array}
\end{equation*}
We define $K \in \br^{n \times n}$ with $K_{ij} = k_{XY}((\tilde{x}_i, \tilde{y}_i),(\tilde{x}_j, \tilde{y}_j))$ and $R$ as an upper triangular matrix for the Cholesky decomposition of $K$. We let $\Phi_i$ be the $i^\textnormal{th}$ column of $R$. Then, the dual problem of Eq.~\eqref{prob:W2-empirical} reads:
\begin{equation}\label{prob:main}
\begin{array}{cl}
\min\limits_{\gamma \in \br^n} & \tfrac{1}{4\lambda_2}\gamma^\top Q\gamma - \tfrac{1}{2\lambda_2}\gamma^\top z + \tfrac{q^2}{4\lambda_2}, \\ 
\st & \sum_{i=1}^n \gamma_i\Phi_i\Phi_i^\top + \lambda_1 I \succeq 0. 
\end{array}
\end{equation}
Suppose that $\hat{\gamma}$ is one such minimizer, we have
\begin{equation*}
\widehat{\OT}^n = \tfrac{q^2}{2\lambda_2} - \tfrac{1}{2\lambda_2}\sum_{i=1}^n \hat{\gamma}_i(w_{\hat{\mu}}(\tilde{x}_i) + w_{\hat{\nu}}(\tilde{y}_i)). 
\end{equation*}
To the best of our knowledge, the only method proposed to solve Eq.~\eqref{prob:main} is the SSIPM, for which the required number of iterations is known to grow as $n$ grows. To avoid this issue,~\citet{Muzellec-2021-Near} proposed solving an unconstrained relaxation model, which allows for the application of gradient-based methods. However, the estimators obtained from solving such relaxations lack any statistical guarantee. 

%!TEX root = main.tex
\section{Method and Analysis}\label{sec:alg}
In this section, we derive our algorithm and provide a convergence rate analysis. We define first a suitable root function that is optimized by kernel-based OT, and apply the regularized SSN method. We improve the computation of each SSN step by exploring the special structure of the generalized Jacobian of that function. We also safeguard the regularized SSN method using a min-max method to achieve a global rate.

\subsection{A nonsmooth equation model for kernel-based OT}
We define the operator $\Phi: \br^{n \times n} \mapsto \br^n$ and its adjoint $\Phi^\star: \br^n \mapsto \br^{n \times n}$ as
\begin{equation*}
\Phi(X) = \begin{pmatrix} \langle X, \Phi_1\Phi_1^T\rangle \\ \vdots \\ \langle X, \Phi_n\Phi_n^T\rangle \end{pmatrix}, \quad \Phi^\star(\gamma) = \sum_{i=1}^n \gamma_i\Phi_i\Phi_i^T. 
\end{equation*}
Clearly, Eq.~\eqref{prob:main} can be reformulated as the following optimization problem given by 
\begin{equation}\label{prob:minimax}
\begin{array}{cl}
\min\limits_{\gamma \in \br^n} \max\limits_{X \in \SCal_+^n} & \tfrac{1}{4\lambda_2}\gamma^T Q\gamma - \tfrac{1}{2\lambda_2}\gamma^T z + \tfrac{q^2}{4\lambda_2} - \langle X, \Phi^\star(\gamma) + \lambda_1 I\rangle. 
\end{array}
\end{equation}
We denote $w = (\gamma, X)$ as a vector-matrix pair and let $R: \br^n \times \br^{n \times n} \rightarrow \br^n \times \br^{n \times n}$ be given by
\begin{equation}\label{def:residue}
R(w) = \begin{pmatrix} \tfrac{1}{2\lambda_2}Q\gamma - \tfrac{1}{2\lambda_2}z - \Phi(X) \\ X - \proj_{\SCal_+^n}(X - (\Phi^\star(\gamma) + \lambda_1 I)) \end{pmatrix}. 
\end{equation}
where $\SCal_+^n = \{X \in \br^{n \times n}: X \succeq 0, X^T = X\}$. Then, we measure the optimality of $w$ by monitoring $\|R(w)\|$, as supported by the following proposition linking $R$ to minimizers of Eq.~\eqref{prob:main}.
\begin{proposition}\label{Prop:equivalence-main}
A point $\hat{\gamma} \in \br^n$ is an optimal solution of Eq.~\eqref{prob:main} if and only if $\hat{w} = (\hat{\gamma}, \hat{X})$ satisfies $R(\hat{w}) = 0$ for some $\hat{X} \in \SCal_+^n$. 
\end{proposition}
Proposition~\ref{Prop:equivalence-main} shows that we can recover a kernel-based OT estimator by solving the nonsmooth equation model $R(w) = 0$.

\paragraph{Regularized SSN method.} Since $R$ is Lipschitz, Rademacher's theorem guarantess that $R$ is almost everywhere differentiable. We introduce generalized Jacobians~\citep{Clarke-1990-Optimization}. 
\begin{definition}\label{def:generalized_subdiff}
Suppose $R$ is Lipschitz and $D_R$ is the set of differentiable points of $R$. The B-subdifferential at $w$ is $\partial_B R(w) := \{\lim_{k \rightarrow +\infty} \nabla F(w^k) \mid w^k \in D_R, w^k \rightarrow w\}$ and the generalized Jacobian at $w$ is $\partial R(w) = \conv(\partial_B R(w))$ where $\conv$ is the convex hull.  
\end{definition}
The regularized SSN method for solving $R(w) = 0$ is as follows: Having the vector $w_k$, we compute $w_{k+1} = w_k + \Delta w_k$ where $\Delta w_k$ is obtained by solving
\begin{equation}\label{scheme:SSN}
(\JCal_k + \mu_k \ICal)[\Delta w_k] = -r_k, 
\end{equation}
where $\JCal_k \in \partial R(w_k)$, $r_k = R(w_k)$ and $\ICal$ is the identity. The parameter is chosen as $\mu_k = \theta_k\|r_k\|$ to stabilize the SSN method in practice. If $R$ is continuously differentiable and $\theta_k = 0$, the regularized SSN method reduces to the classical regularized Newton method which attains a local quadratic rate. Although the regularized SSN method is divergent in general~\citep{Kummer-1988-Newton}, its local superlinear rate has been proved if $R$ is strongly semi-smooth~\citep{Qi-1993-Nonsmooth, Zhou-2005-Superlinear, Xiao-2018-Regularized}. 

\subsection{Properties of the nonsmooth map $R$}
\paragraph{Generalized Jacobian.} Let us focus on the structure of the generalized Jacobian of $R(w)$. Using the definition of $\SCal_+^n$, one has $\proj_{\SCal_+^n}(Z) = P_\alpha \Sigma_\alpha P_\alpha^T$
where 
\begin{equation}\label{def:SD}
Z = P\Sigma P^T = \begin{pmatrix} P_\alpha & P_{\bar{\alpha}} \end{pmatrix} \begin{pmatrix} \Sigma_\alpha & 0 \\ 0 & \Sigma_{\bar{\alpha}} \end{pmatrix} \begin{pmatrix} P_\alpha^T \\ P_{\bar{\alpha}}^T \end{pmatrix}\,,
\end{equation}
with $\Sigma = \diag(\sigma_1, \ldots, \sigma_n)$, with the sets of indices of positive and nonpositive eigenvalues of $Z$  written $\alpha = \{i \mid \sigma_i > 0\}$ and $\bar{\alpha} = \{1, \ldots, n\} \setminus \alpha$. 

We define a generalized operator $\MCal(Z) \in \partial \proj_{\SCal_+^n}(Z)$ using its application to an $n \times n$ matrix $S$: 
\begin{equation*}
\MCal(Z)[S] = P(\Omega \circ (P^T SP))P^T \textnormal{ for all } S \in \SCal_+^n, 
\end{equation*}
where the $\circ$ symbol denotes a Hadamard product and $\Omega = \begin{pmatrix} E_{\alpha\alpha} & \eta_{\alpha\bar{\alpha}} \\ \eta_{\alpha\bar{\alpha}}^T & 0 \end{pmatrix}$ with $E_{\alpha\alpha}$ being a matrix of ones and $\eta_{ij} = \frac{\sigma_i}{\sigma_i - \sigma_j}$ for all $(i, j) \in \alpha \times \bar{\alpha}$. Note that all entries of $\Omega$ lie in the interval $(0, 1]$. In general, it is nontrivial to characterize the generalized Jacobian $\partial R(w)$ exactly but we can compute an element $\JCal(w) \in \partial R(w)$ using $\MCal(\cdot)$ as defined before.  

We next introduce the definition of the (strong) semismoothness of an operator. 
\begin{definition}\label{def:semismooth}
Suppose that $R$ is Lipschitz, we say it is (strongly) semismooth at $w$ if (i) $R$ is directionally differentiable at $w$; and (ii) for any $\JCal \in \partial R(w + \Delta w)$, we let $\Delta w \rightarrow 0$ and have
\begin{equation*}
\begin{array}{rl}
\textnormal{(semismooth)} & \tfrac{\|R(w + \Delta w) - R(w) - \JCal[\Delta w]\|}{\|\Delta w\|} \rightarrow 0, \\ \textnormal{(strongly semismooth)} & \tfrac{\|R(w + \Delta w) - R(w) - \JCal[\Delta w]\|}{\|\Delta w\|^2} \leq C.  \end{array}
\end{equation*}
\end{definition}
The following proposition characterizes the residual map given in Eq.~\eqref{def:residue} and guarantees that the SSN method is suitable to solve $R(w) = 0$. 
\begin{proposition}\label{Prop:strong-semismooth}
The residual map $R$ in Eq.~\eqref{def:residue} is strongly semismooth. 
\end{proposition}

\subsection{Newton updates}
We discuss how to compute the Newton direction $\Delta w_k$ efficiently. From a computational point of view, it is not practical to solve the linear system in Eq.~\eqref{scheme:SSN} exactly. Thus, we seek an approximation step $\Delta w_k$ by solving Eq.~\eqref{scheme:SSN} approximately such that 
\begin{equation}\label{criterion:SSN_inexact}
\|(\JCal_k + \mu_k\ICal)[\Delta w_k] + r_k\| \leq \tau\min\{1, \kappa\|r_k\|\|\Delta w_k\|\}, 
\end{equation}
where $0 < \tau, \kappa < 1$ are some positive constants and $\|\cdot\|$ is defined for a vector-matrix pair $w = (\gamma, X)$ (i.e., $\|w\| = \|\gamma\|_2 + \|X\|_F$ where $\|\cdot\|_2$ is Euclidean norm and $\|\cdot\|_F$ is Frobenius norm). 
\begin{algorithm}[!t]
\begin{algorithmic}[1]\caption{Solving Eq.~\eqref{scheme:SSN} where $r_k = (r^1_k, r^2_k)$)}\label{alg:SSN_direction}
\STATE $a^1 = -r^1_k - \frac{1}{\mu_k + 1}(\Phi(r^2_k + \TCal_k[r^2_k]))$ and $a^2 = -r^2_k$. 
\STATE Solve $(\frac{1}{2\lambda_2}\QCal + \mu_k\ICal + \Phi\TCal_k\Phi^\star)^{-1}\tilde{a}^1 = a^1$ inexactly and compute $\tilde{a}^2 = \tfrac{1}{\mu_k + 1}(a^2 + \TCal_k[a^2])$, where $\TCal_k[\cdot]$ is computed using the trick~\citep{Zhao-2010-Newton}. 
\STATE Compute the direction $\Delta w_k = (\Delta w_k^1, \Delta w_k^2)$ by $\Delta w_k^1 = \tilde{a}^1$ and $\Delta w_k^2 = \tilde{a}^2 - \TCal_k[\Phi^\star(\tilde{a}^1)]$. 
\end{algorithmic}
\end{algorithm}
Since $\JCal_k$ in Eq.~\eqref{scheme:SSN} is nonsymmetric and its dimension is large, we use the Schur complement formula to transform Eq.~\eqref{scheme:SSN} into a smaller symmetric system. If we vectorize the vector-matrix pair\footnote{If $w = (\gamma, X)$ is a vector-matrix pair, we define $\textnormal{vec}(w) = (\gamma; \textnormal{vec}(X))$ as its vectorization.} $\Delta w$, the operators $\MCal(Z)$ and $\Phi$ can be expressed as matrices: 
\begin{equation*}
M(Z) = \tilde{P}\Gamma\tilde{P}^T \in \br^{n^2 \times n^2}, \; A = \begin{pmatrix} \Phi_1^T \otimes \Phi_1^T \\ \vdots \\ \Phi_n^T \otimes \Phi_n^T \end{pmatrix} \in \br^{n \times n^2}, 
\end{equation*}
where $\tilde{P} = P \otimes P$ and $\Gamma = \diag(\textnormal{vec}(\Omega))$. 

We next provide a key lemma on the matrix form of $\JCal_k + \mu_k I$ at a given iterate $w_k = (\gamma_k, X_k)$. 
\begin{lemma}\label{Lemma:key-structure}
Given $w_k = (\gamma_k, X_k)$, we compute $Z_k = X_k - (\Phi^\star(\gamma_k) + \lambda_1 I)$ and use Eq.~\eqref{def:SD} to obtain $P_k$, $\Sigma_k$, $\alpha_k$ and $\bar{\alpha}_k$. We then obtain $\Omega_k$, $\tilde{P}_k = P_k \otimes P_k$ and $\Gamma_k = \diag(\textnormal{vec}(\Omega_k))$. Then, the matrix form of $\JCal_k + \mu_k I$ is given by 
\begin{equation*}
\begin{aligned}
(J_k &+ \mu_k I)^{-1} = C_1 BC_2, \text{ where }\\
C_1 &= \begin{pmatrix} I & 0 \\ -T_k A^T & I \end{pmatrix}, \quad C_2 = \begin{pmatrix} I & \tfrac{1}{\mu_k + 1}(A + AT_k) \\ 0 & I \end{pmatrix},\\
B &= \begin{pmatrix} (\tfrac{1}{2\lambda_2}Q + \mu_k I + AT_kA^T)^{-1} & 0 \\ 0 & \tfrac{1}{\mu_k + 1}(I + T_k) \end{pmatrix}, 
\end{aligned}
\end{equation*}
with $T_k = \tilde{P}_k L_k\tilde{P}_k^T$ where $L_k$ is a diagonal matrix with $(L_k)_{ii} = \frac{(\Gamma_k){ii}}{\mu_k + 1 - (\Gamma_k)_{ii}}$ and $(\Gamma_k)_{ii} \in (0, 1]$ is then denoted as the $i^\textnormal{th}$ diagonal entry of $\Gamma_k$. 
\end{lemma}
As a consequence of Lemma~\ref{Lemma:key-structure}, the solution of Eq.~\eqref{scheme:SSN} can be obtained by solving one certain symmetric linear system with the matrix $\tfrac{1}{2\lambda_2}Q + \mu_k I + AT_kA^T$. We remark that this system is well-defined since both $Q$ and $AT_kA^T$ are positive semidefinite and the coefficient $\mu_k$ is chosen such that $\tfrac{1}{2\lambda_2}Q + \mu_k I + AT_kA^T$ is invertible. This also shows that Eq.~\eqref{scheme:SSN} is well-defined. 

We define $\TCal_k$ and $\QCal$ as the operator form of $T_k = \tilde{P}_kL_k\tilde{P}_k^T$ and $Q$ and write $r_k = (r^1_k, r^2_k)$ explicitly where $r^1_k \in \br^n$ and $r^2_k \in \br^{n \times n}$. Then, we have
\begin{equation*}
\textnormal{vec}(a) = -\begin{pmatrix} I & \tfrac{1}{\mu_k + 1}(A + AT) \\ 0 & I \end{pmatrix}\textnormal{vec}(r_k) \Longrightarrow \left\{\begin{array}{l}
a^1 = -r^1_k - \tfrac{1}{\mu_k + 1}(\Phi(r^2_k + \TCal_k[r^2_k])), \\ a^2 = -r^2_k.
\end{array}\right.
\end{equation*}
The next step consists in solving a new symmetric linear system and is given by 
\begin{equation*}
\textnormal{vec}(\tilde{a}) = \begin{pmatrix} \left(\tfrac{Q}{2\lambda_2} + \mu_k I + AT_k A^T\right)^{-1}\!\!\!\!\! & 0 \\ 0 & \!\!\!\!\!\!\!\tfrac{1}{1+\mu_k}(I + T_k) \end{pmatrix}\textnormal{vec}(a), 
\end{equation*}
which leads to 
\begin{equation*}
\left\{\begin{array}{l}
\tilde{a}^1 = (\tfrac{1}{2\lambda_2}\QCal + \mu_k\ICal + \Phi\TCal_k\Phi^\star)^{-1}a^1, \\ \tilde{a}^2 = \tfrac{1}{\mu_k + 1}(a^2 + \TCal_k[a^2]).\end{array}\right.
\end{equation*}
Compared to Eq.~\eqref{scheme:SSN} whose matrix form has size $(n^2 + n) \times (n^2 + n)$, we remark that the one in the step above is smaller with the size of $n \times n$ and can be efficiently solved using conjugate gradient (CG) or symmetric quasi-minimal residual (QMR) methods~\citep{Kelley-1995-Iterative, Saad-2003-Iterative}. The final step is to compute the Newton direction $\Delta w_k = (\Delta w_k^1, \Delta w_k^2)$ as follows, 
\begin{equation*}
\textnormal{vec}(\Delta w_k) = \begin{pmatrix} I & 0 \\ -TA^T & I \end{pmatrix}\textnormal{vec}(\tilde{a}) \Longrightarrow  \left\{\begin{array}{l} \Delta w_k^1 = \tilde{a}^1, \\ \Delta w_k^2 = \tilde{a}^2 - \TCal_k[\Phi^\star(\tilde{a}^1)].
\end{array}\right.
\end{equation*}
It remains to provide an efficient manner to compute $\TCal_k[\cdot]$. Since $\TCal_k$ is defined as the operator form of $T = \tilde{P}_k L_k\tilde{P}_k^T$, we have
\begin{equation*}
\TCal_k[S] = P_k(\Psi_k \circ (P_k^T SP_k))P_k^T, 
\end{equation*}
where $\Psi_k$ is determined by $\mu_k$ and $\Omega_k$: Indeed,
\begin{equation*}
\Omega_k = \begin{pmatrix} E_{\alpha_k\alpha_k} & \eta_{\alpha_k\bar{\alpha}_k} \\ \eta_{\alpha_k\bar{\alpha}_k}^T & 0 \end{pmatrix} \Longrightarrow \Psi_k = \begin{pmatrix} \tfrac{1}{\mu_k}E_{\alpha_k\alpha_k} & \xi_{\alpha_k\bar{\alpha}_k} \\ \xi_{\alpha_k\bar{\alpha}_k}^T & 0 \end{pmatrix}, 
\end{equation*}
where we have $\xi_{ij} = \tfrac{\eta_{ij}}{\mu_k + 1 - \eta_{ij}}$ for any $(i, j) \in \alpha_k \times \bar{\alpha}_k$. Following~\citet{Zhao-2010-Newton}, we use the decomposition $\TCal_k[S] = G + G^T$ where $U = P_k(:, \alpha_k)^T S$ and  
\begin{equation*}
G = P_k(:, \alpha_k)(\tfrac{1}{2\mu_k}(UP_k(:, \alpha_k))P_k(:, \alpha_k)^\top + \xi_{\alpha_k\bar{\alpha}_k} \circ (UP_k(:, \bar{\alpha}_k))P_k(:, \bar{\alpha}_k)^\top). 
\end{equation*} 
The number of flops for computing $\TCal_k[S]$ is $8|\alpha_k|n^2$. If $|\alpha_k| > \bar{\alpha}_k$, we use $\TCal_k[S] = \frac{1}{\mu_k}S - P_k((\frac{1}{\mu_k} E - \Psi_k) \circ (P_k^\top SP_k))P_k^\top$ with $8|\bar{\alpha}_k|n^2$ flops. Thus, we obtain an inexact solution of Eq.~\eqref{scheme:SSN} efficiently whenever $|\alpha_k|$ or $|\bar{\alpha}_k|$ is small. We present the scheme for computing an inexact SSN direction in Algorithm~\ref{alg:SSN_direction}. 
\begin{algorithm}[!t]
\begin{algorithmic}[1]\caption{Our specialized SSN method}\label{alg:main}
\STATE \textbf{Input:} $\tau, \kappa$, $\alpha_2 \geq \alpha_1 > 0$, $\beta_0, \beta_1 < 1$, $\beta_2 > 1$ and $\underline{\theta}, \overline{\theta} > 0$. 
\STATE \textbf{Initialization:} $v_0 = w_0 \in \br^n \times \SCal_+^n$ and $\theta_0 > 0$. 
\FOR{$k = 0, 1, 2, \ldots$}
\STATE Update $v_{k+1}$ from $v_k$ using one-step EG.  
\STATE Select $\JCal_k \in \partial R(w_k)$. 
\STATE Solve the linear system in Eq.~\eqref{scheme:SSN} approximately such that $\Delta w_k$ satisfies Eq.~\eqref{criterion:SSN_inexact}. 
\STATE Compute $\tilde{w}_{k+1} = w_k + \Delta w_k$. 
\STATE Update $\theta_{k+1}$ in the adaptive manner. 
\STATE Set $w_{k+1} = \tilde{w}_{k+1}$ if $\|R(\tilde{w}_{k+1})\| \leq \|R(v_{k+1})\|$ is satisfied. Otherwise, set $w_{k+1} = v_{k+1}$. 
\ENDFOR
\end{algorithmic}
\end{algorithm}
We propose a rule for updating $\theta_k$ where $\mu_k = \theta_k\|r_k\|$ will be used in Eq.~\eqref{scheme:SSN}. It is summarized as follows: 
\begin{equation*}
\theta_{k+1} = \left\{
\begin{array}{cl}
\max\{\underline{\theta}, \beta_0\theta_k\}, & \textnormal{if } \rho_k \geq \alpha_2\|\Delta w_k\|^2, \\
\beta_1\theta_k, & \textnormal{if } \alpha_1\|\Delta w_k\|^2 \leq \rho_k < \alpha_2\|\Delta w_k\|^2, \\
\min\{\overline{\theta}, \beta_2\theta_k\}, & \textnormal{otherwise}. 
\end{array}
\right. 
\end{equation*}
where $\rho_k = -\langle R(w_k + \Delta w_k), \Delta w_k\rangle$, $\beta_0 < 1, \beta_1, \beta_2 > 1$ and $\underline{\theta}, \overline{\theta} > 0$. Intuitively, $\theta_k$ can control the quality of $\Delta w_k$ and the larger value of $\theta_k$ gives a slow yet stable convergence. The small value of $\frac{\rho_k}{\|\Delta w_k\|^2}$ implies that $\Delta w_k$ is bad and we shall increase $\theta_k$.
\begin{remark}
We see that the per-iteration cost is significantly reduced since we have shown that solving the linear system in Eq.~\eqref{scheme:SSN} whose matrix form has size $(n^2 + n) \times (n^2 + n)$ can be equivalently reduced to solving a much smaller linear system whose matrix form has size $n \times n$. Such equivalent reduction is based on Lemma~\ref{Lemma:key-structure} whose proof is summarized in Appendix~\ref{app:key-structure}. 
\end{remark}

\subsection{Algorithm}
We summarize our approach in Algorithm~\ref{alg:main}. We generate a sequence of iterates by alternating between extragradient (EG) method~\citep{Facchinei-2007-Finite, Cai-2022-Finite} and the regularized SSN method.  

We maintain an auxiliary sequence of iterates $\{v_k\}_{k \geq 0}$. This sequence is directly generated by the EG method for solving the min-max problem in Eq.~\eqref{prob:minimax} and is used to safeguard the regularized SSN method to achieve a global convergence rate. In particular, we start with $v_0 = w_0 \in \br^n \times \SCal_+^n$ and perform the $k^{\textnormal{th}}$ iteration as follows, 
\begin{enumerate}
\item Update $v_{k+1}$ from $v_k$ via 1-step EG.  
\item Update $\tilde{w}_{k+1}$ from $w_k$ via 1-step regularized SSN. 
\item Set $w_{k+1} = \tilde{w}_{k+1}$ if $\|R(\tilde{w}_{k+1})\| \leq \|R(v_{k+1})\|$ and $w_{k+1} = v_{k+1}$ otherwise. 
\end{enumerate}
\begin{remark}
The per-iteration cost would be $O(n^3)$ at worst case but it can be much cheaper in practice. Indeed, the $O(n^3)$ cost comes from exactly solving the $n \times n$ linear system. In our experiment, we use CG to approximately solve this linear system and set the maximum iteration number as 20. We can see that the average number of CG steps is less than 5. Also, our implementation can be improved by exploring the potentially sparsity of $Q$, $A$ and $T_k$. In contrast, the linear system at each IPM step becomes severely ill-conditioned as the barrier parameter decreases and the matrix factorization has to be done exactly to achieve high precision. Therefore, our method suffers from the same per-iteration cost as IPM at worst case but can be more flexible and efficient from a practical viewpoint.
\end{remark}
\begin{remark}
Although computing such auxiliary sequence results in extra cost, we can argue that it is not an issue in both theory and practice. Indeed, Theorem~\ref{Thm:convergence-local} guarantees the existence of a local region where 1-step regularized SSN can reduce the residue norm at a quadratic rate. This implies that $\|R(\tilde{w}_{k+1})\| \leq \|R(v_{k+1})\|$ will always hold when $k$ is sufficiently large and $w_{k+1} = v_{k+1}$ will not never hold. This encourages us to stop computing the auxiliary sequence after the iterates enter the local region and only perform the regularized SSN steps. In our experiment, we also find that the iterates are mostly generated by regularized SSN steps. However, it is tricky to implement such strategy since it is hard to check if the generated iterates enter the local region. If we stop computing such auxiliary sequence too early, our algorithm is likely to diverge. To show the power of regularized SSN steps, we also compare our algorithm with pure EG steps in Appendix~\ref{sec:exp-appendix} (see Figure~\ref{fig:syntheticdata-appendix}). 
\end{remark}

\subsection{Convergence Analysis}
We establish the convergence guarantee of Algorithm~\ref{alg:main} in the following theorems. 
\begin{theorem}\label{Thm:convergence-global}
Suppose that $\{w_k\}_{k \geq 0}$ is a sequence of iterates generated by Algorithm~\ref{alg:main}. Then, the residuals of $\{w_k\}_{k \geq 0}$ converge to 0 at a rate of $1/\sqrt{k}$, i.e., $\|R(w_k)\| = O(1/\sqrt{k})$. 
\end{theorem}
In the context of constrained convex-concave min-max optimization,~\citet{Cai-2022-Finite} has proved the $O(1/\sqrt{k})$ last-iterate convergence rate of the EG, matching the lower bounds~\citep{Golowich-2020-Last, Golowich-2020-Tight}. Since the kernel-based OT estimation can be solved as a min-max problem, the global convergence rate  in Theorem~\ref{Thm:convergence-global} demonstrates the efficiency of Algorithm~\ref{alg:main}. It remains unclear whether or not we can improve these results by exploring special structure of Eq.~\eqref{prob:minimax}. 

Moreover, such global rate depends on the smoothness parameter of Eq.~\eqref{prob:minimax} rather than the condition number of original formulation of Eq.~\eqref{prob:main}. The explicit dependence on $\lambda_1$ and $\lambda_2$ is unknown since the results of~\citet{Cai-2022-Finite} does not provide the dependence on these problem parameters. Yet, our experiment has shown that our method behaves well when the sample size is medium ($\sim$1000) which is sufficient for kernel-based OT estimation in most cases. 
\begin{theorem}\label{Thm:convergence-local}
Suppose that $\{w_k\}_{k \geq 0}$ is a sequence of iterates generated by Algorithm~\ref{alg:main}. Then, the residual norm at $\{w_k\}_{k \geq 0}$ converge to 0 at a quadratic rate, i.e., $\|R(w_{k+1})\| \leq C\|R(w_k)\|^2$ for some constant $C > 0$, if the initial point $w_0$ is sufficiently close to $w^\star$ with $R(w^\star) = 0$ and every element of $\partial R(w^\star)$ is invertible. 
\end{theorem}
Similar to classical Newton methods which are key ingredients for IPM, the regularized SSN methods enjoy the weak dependence on problem conditioning; see~\citet{Qi-1993-Nonsmooth} for the details. 
\begin{remark}
Our algorithm becomes inefficient when $\epsilon$ is small but has better dependence on $n$ than IPM. This is more desirable since the large $n$ is necessary to ensure good statistical approximation (see~\citet[Page 11-12]{Muzellec-2021-Near} for details). Such trade-off between $n$ and $1/\epsilon$ has occurred in the computation of plug-in estimators: despite worse dependence on $1/\epsilon$, the Sinkhorn method is recognized as more efficient than IPM in practice since many applications require low-accurate solution ($\epsilon \sim 10^{-2}$) when the sample size $n$ is large. In addition, we remark that our algorithm does not downgrade the value of IPM since the latter one is more suitable when $\epsilon$ is small. 
\end{remark}

\begin{figure*}[!t]\centering
\includegraphics[width=.3\textwidth]{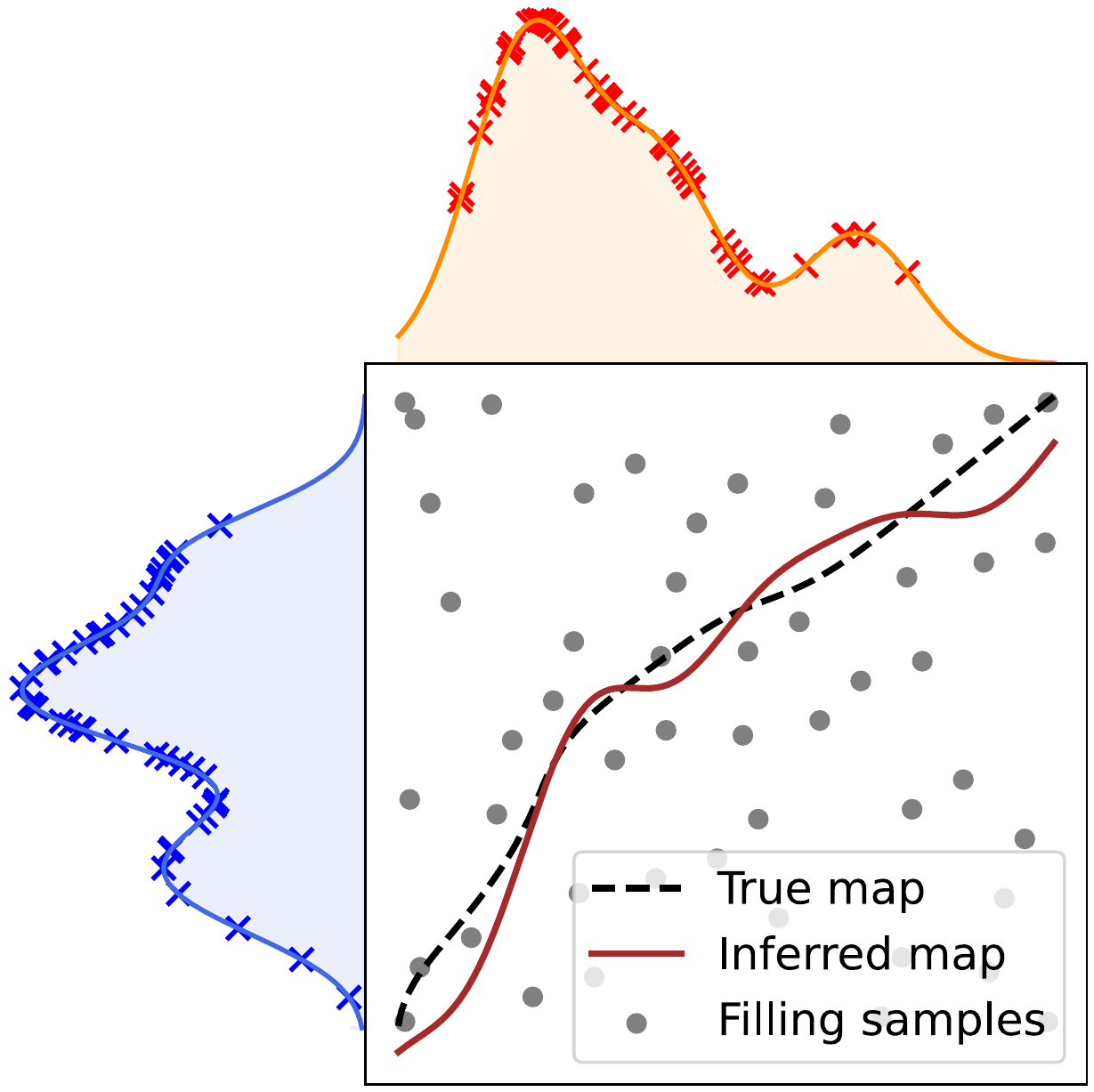}
\includegraphics[width=.3\textwidth]{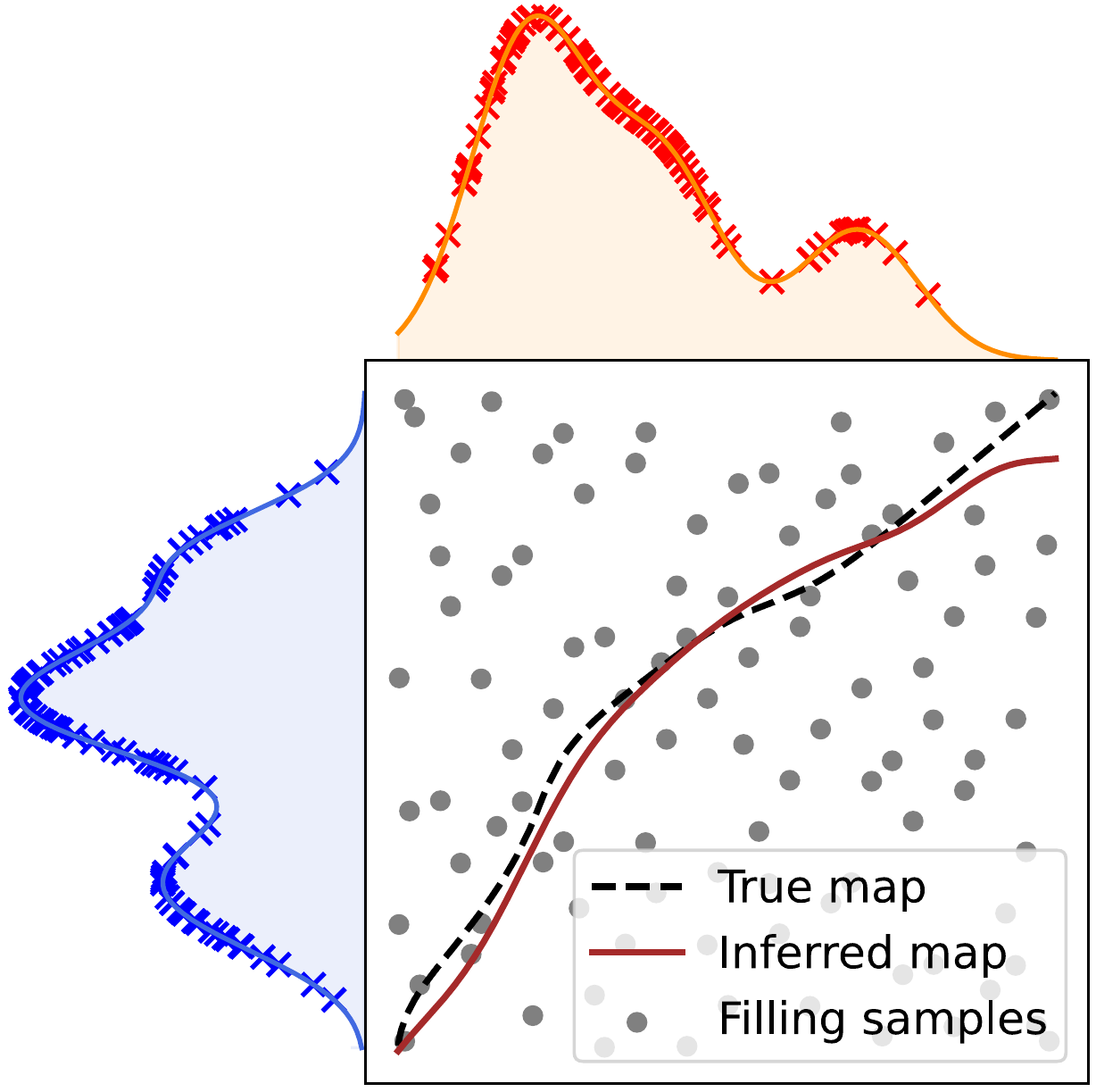}
\includegraphics[width=.3\textwidth]{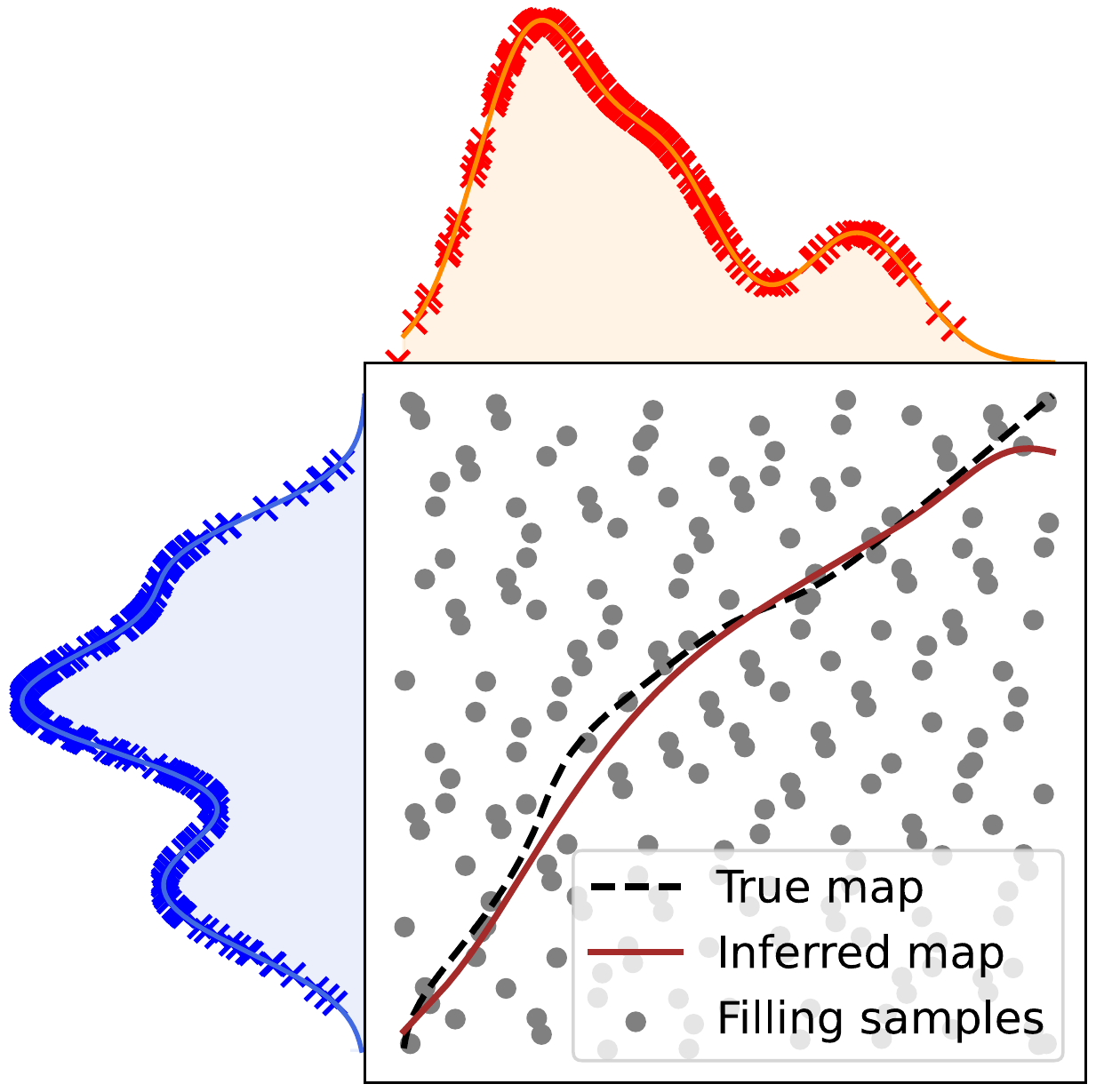}
\caption{Visualization of the OT map with $n_{\textnormal{sample}} = n \in \{50, 100 ,200\}$. } \vspace*{-.5em}\label{fig:synthetic-map}
\end{figure*}
\begin{figure*}[!t]\centering
\includegraphics[width=.3\textwidth]{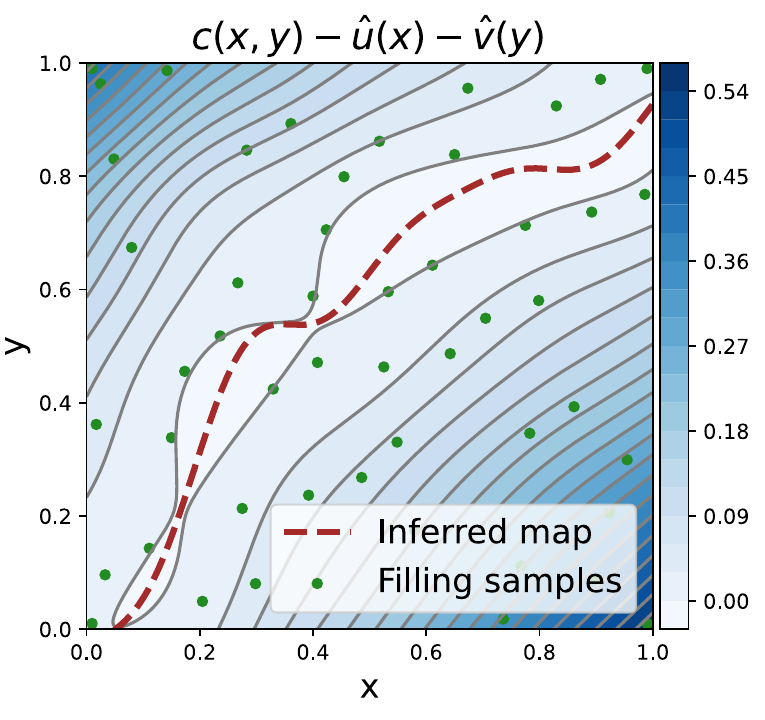}
\includegraphics[width=.3\textwidth]{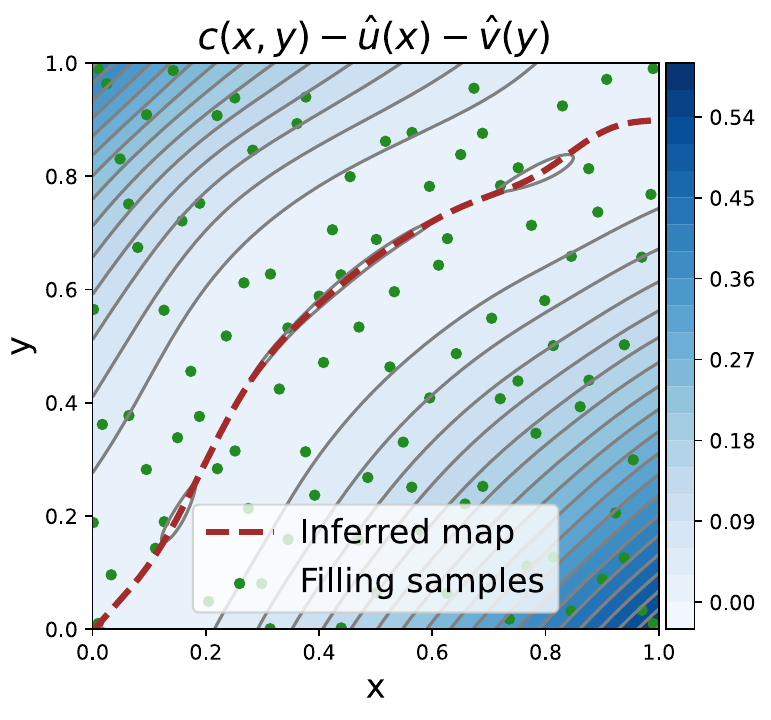}
\includegraphics[width=.3\textwidth]{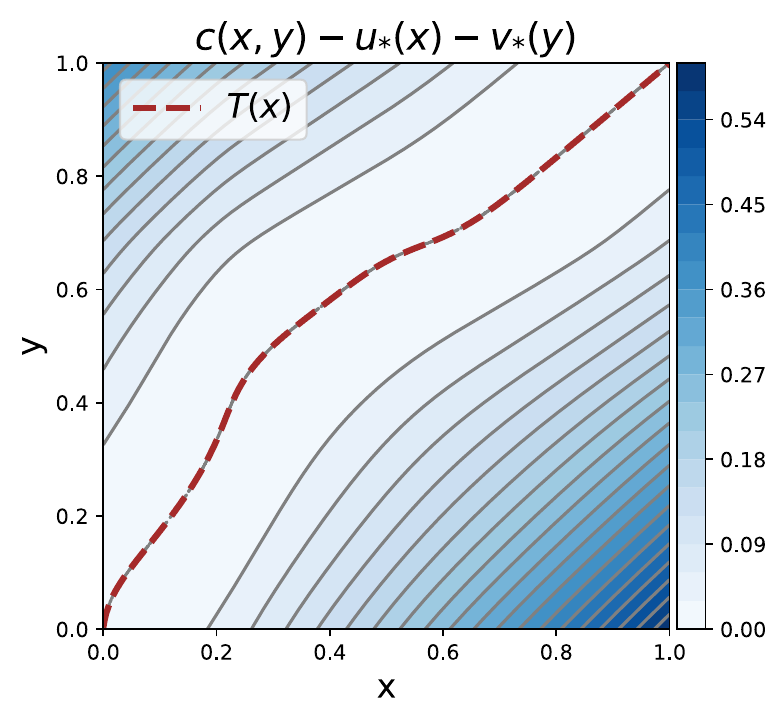}
\caption{Visualization of the constraint: \textit{(left, middle)} $n_{\textnormal{sample}} = n \in \{50, 100\}$, \textit{(right)} ground truth.} \vspace*{-.5em}
\label{fig:synthetic-constraint}
\end{figure*}
\section{Experiments}\label{sec:exp}
We present several experimental results for computing kernel-based OT estimators run with our SSN algorithm. The baseline approach here is the SSIPM~\citep{Vacher-2021-Dimension}; we exclude the gradient-based method~\citep{Muzellec-2021-Near} from our experiment since it is designed to solve a different relaxation model. All experiments were conducted on a MacBook Pro with an Intel Core i9 2.4GHz and 16GB memory. For Algorithm~\ref{alg:main}, we set $\alpha_1=10^{-6}$, $\alpha_2=1.0$, $\beta_0=0.5$, $\beta_1=1.2$ and $\beta_2=5$. 

\paragraph{SSIPM vs SSN on Synthetic data.} Following the setup in~\citet{Vacher-2021-Dimension}, we draw $n_\textnormal{sample}$ samples from $\mu$ and $n_\textnormal{sample}$ samples from $\nu$, where $\mu$ is a mixture of 3 $d$-dimensional Gaussian distributions and $\nu$ is a mixture of 5 $d$-dimensional Gaussian distributions. Then, we sample $n$ filling samples from a $2d$ Sobol sequence. We also set the bandwidth $\sigma^2 = 0.005$ and parameters $\lambda_1 = \frac{1}{n}$ and $\lambda_2 = \frac{1}{\sqrt{n_\textnormal{sample}}}$. Focusing on 1-dimensional setting, we report the visualization results in Figure~\ref{fig:synthetic-map} and~\ref{fig:synthetic-constraint} and verify that the inferred OT map gets closer to the true OT map as the number of filling points and data samples increase. 

\begin{figure*}[!t]
\centering
\includegraphics[width=.32\textwidth]{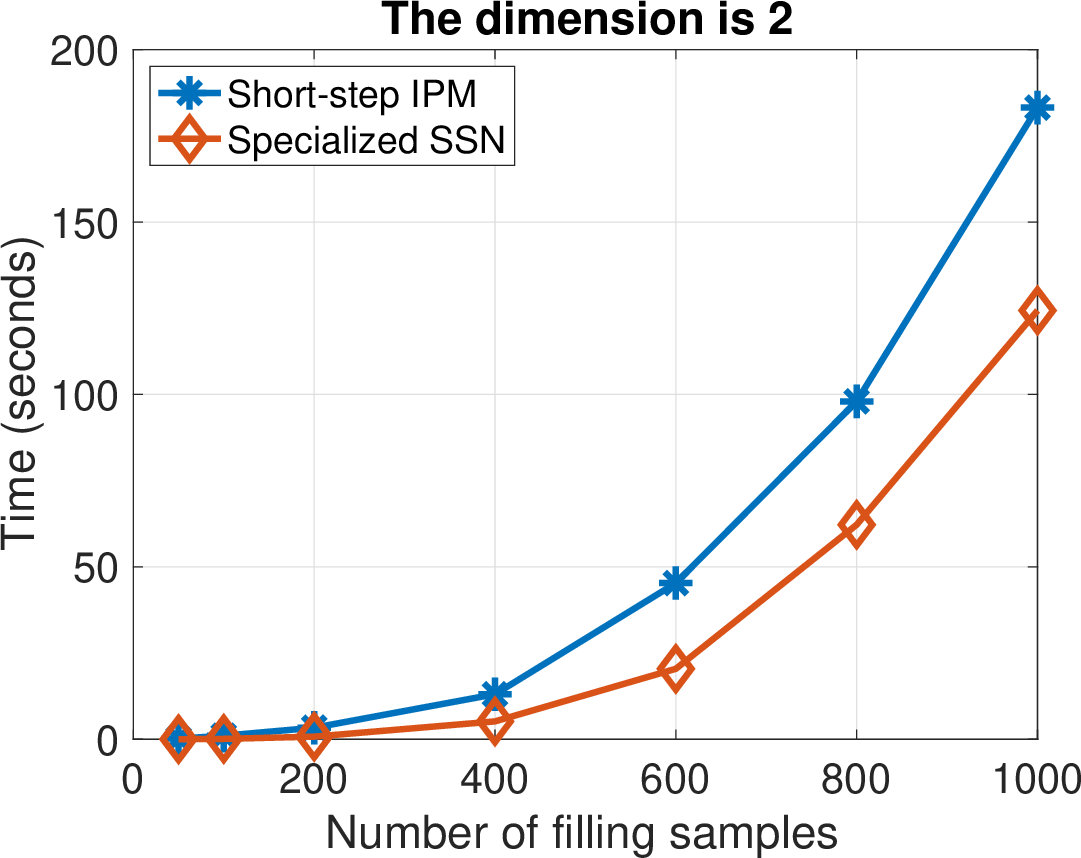} 
\includegraphics[width=.32\textwidth]{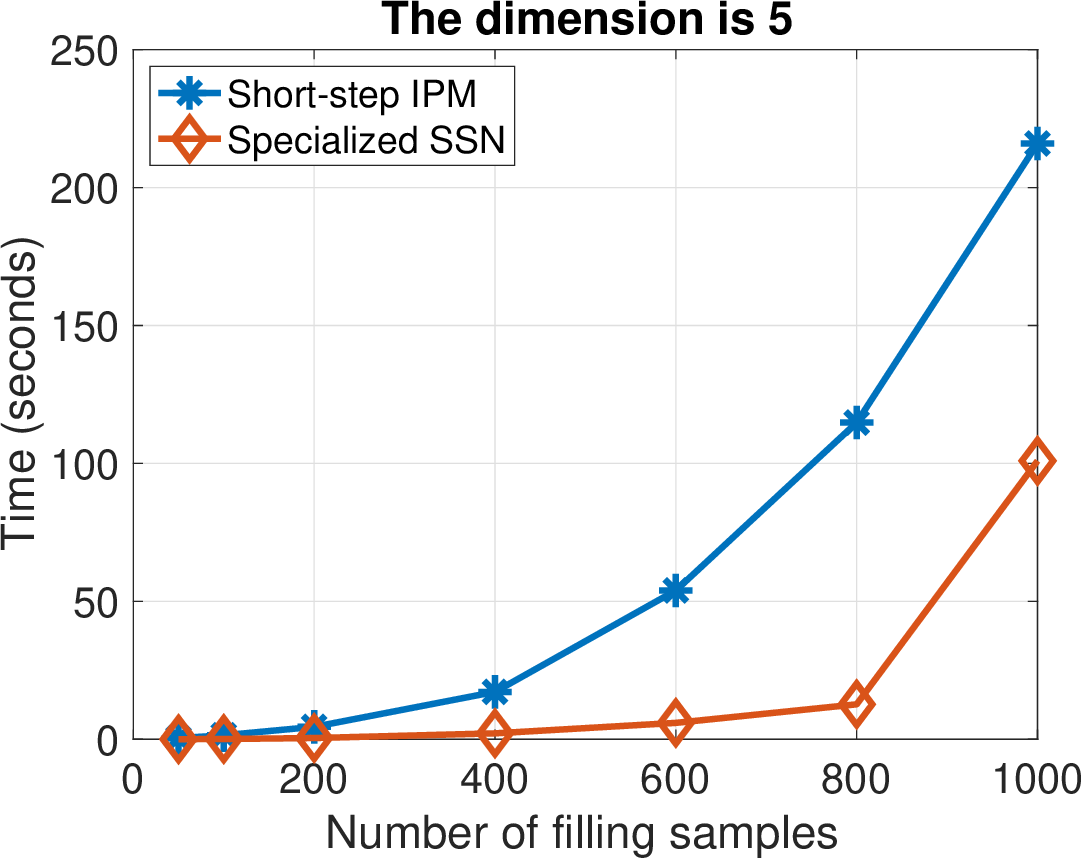} 
\includegraphics[width=.32\textwidth]{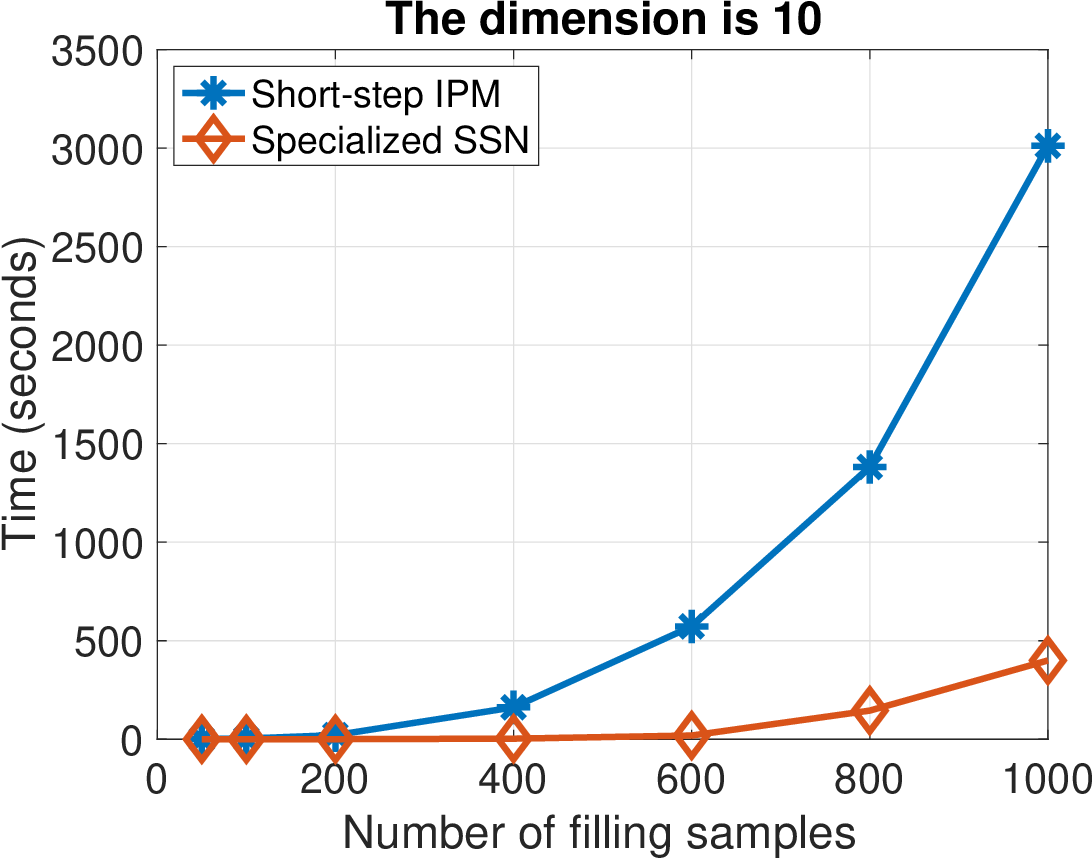} 
\caption{Comparisons of mean computation time of IPM vs. our algorithm (SSN) on CPU time.}\vspace*{-.5em} \label{fig:time}
\end{figure*}

By varying the dimension $d \in \{2, 5, 10\}$, we report the computation efficiency results in Figure~\ref{fig:time}. It indicates that the our new algorithm is more efficient than the IPM as the number of filling points increases, with smaller variance in computation time (seconds). Here, we used the residue norm $\|R(w)\|$ as the measurement and terminated IPM and our method when $\|R(w)\|$ is below than the threshold $0.005$. Although our method can scale to the case of 1000 samples which is relatively small compared to entropic OT methods, these results do start to open up some possibilities. 
\begin{figure*}[!t]
\centering
\includegraphics[width=.32\textwidth]{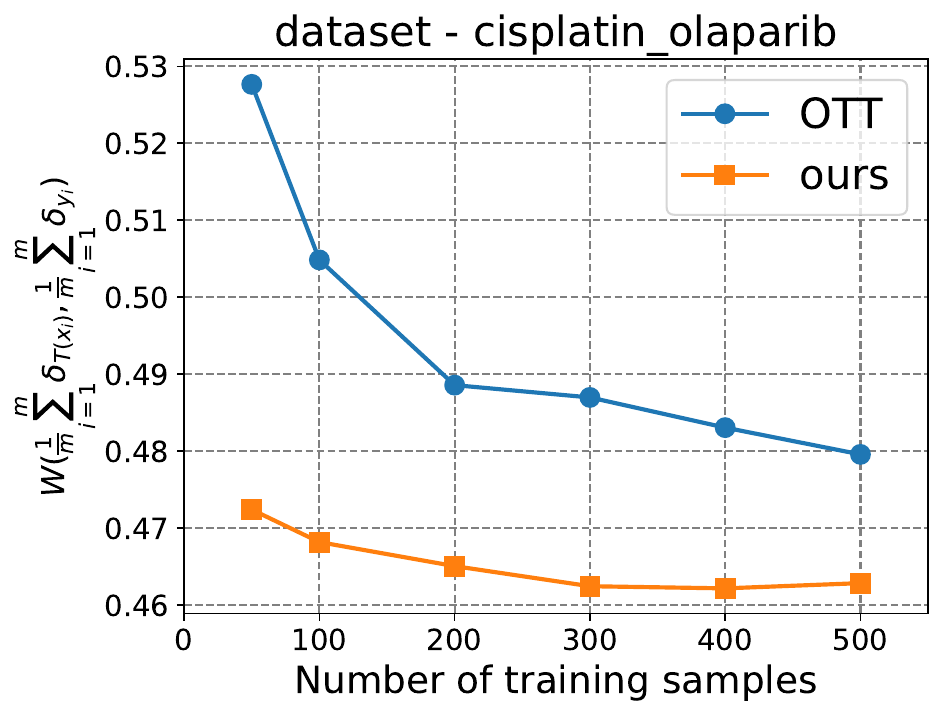}
\includegraphics[width=.32\textwidth]{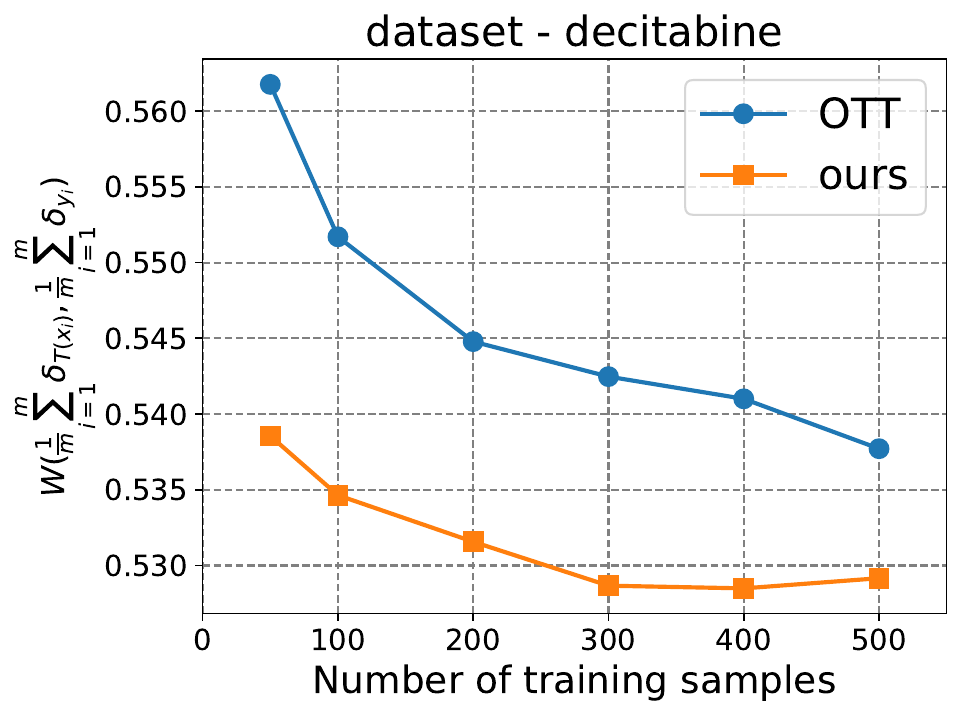}
\includegraphics[width=.32\textwidth]{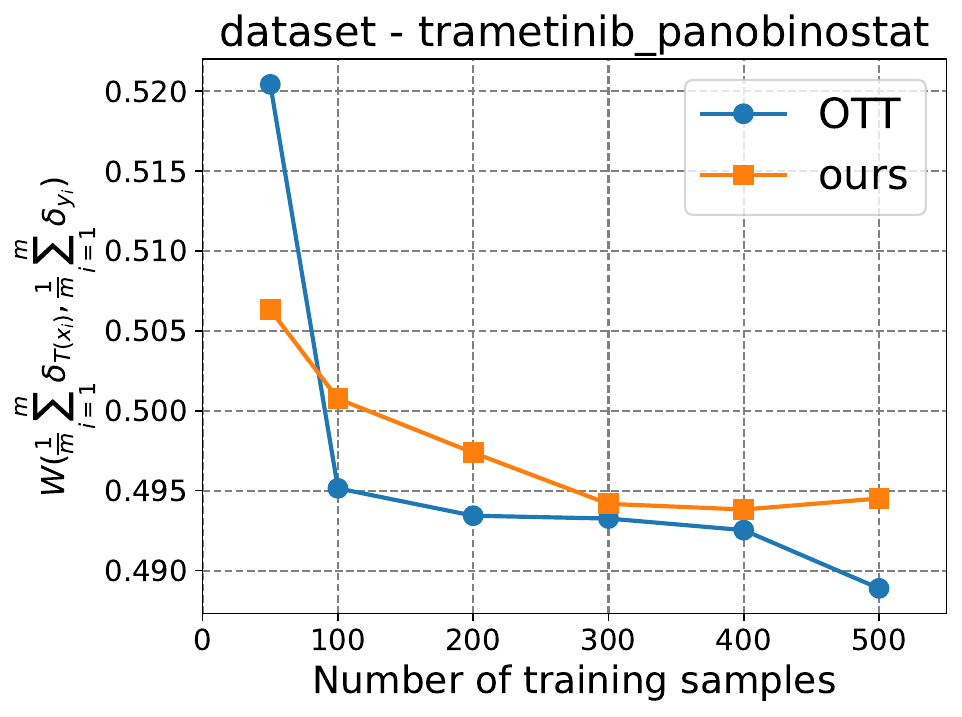}
\includegraphics[width=.32\textwidth]{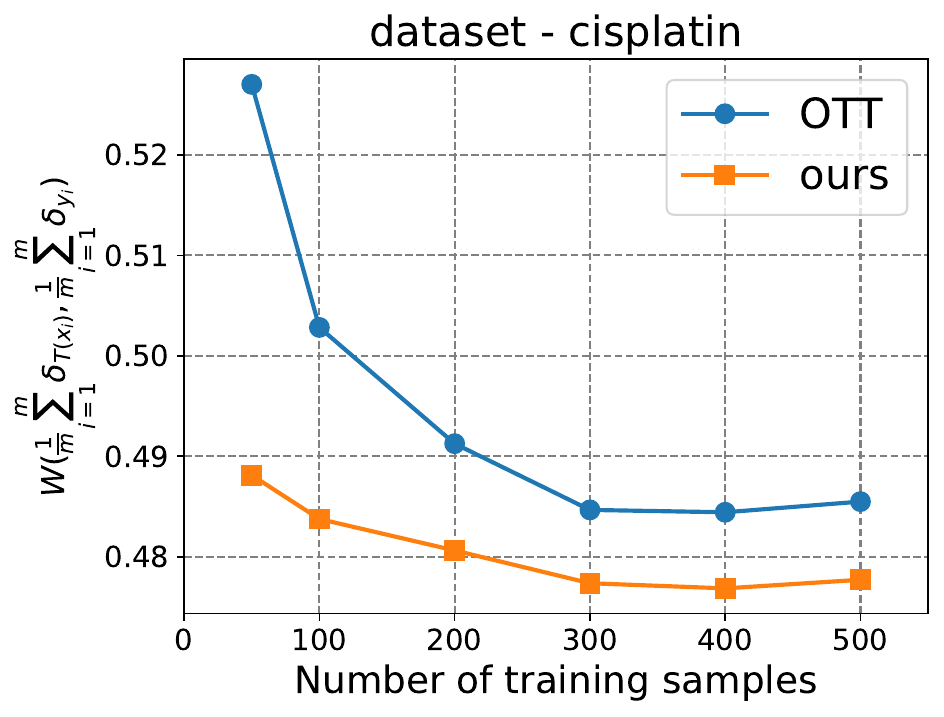}
\includegraphics[width=.32\textwidth]{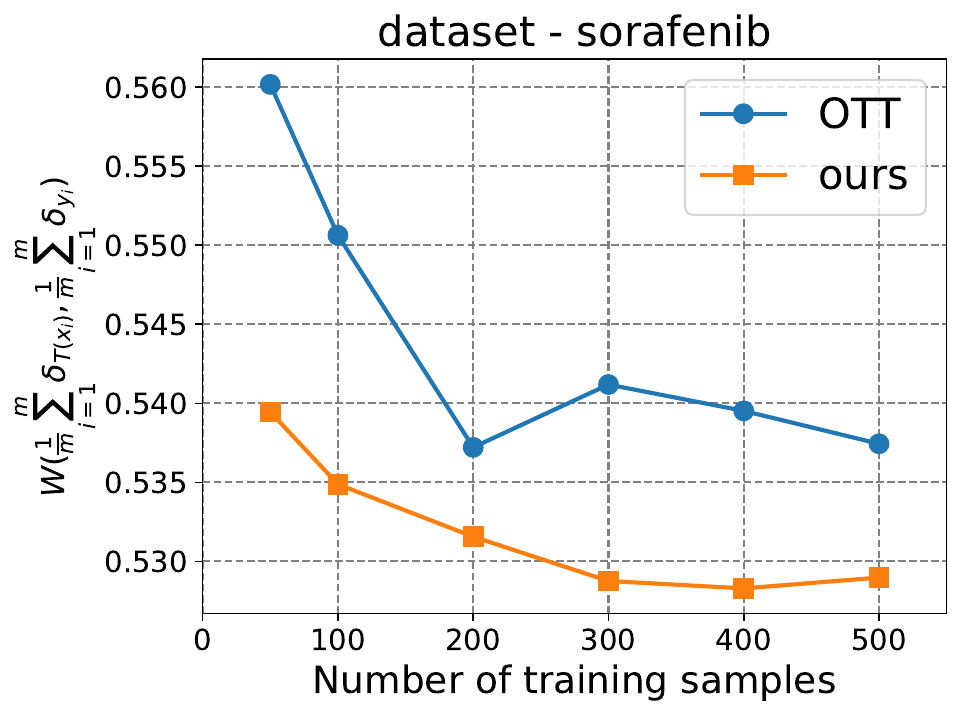}
\includegraphics[width=.32\textwidth]{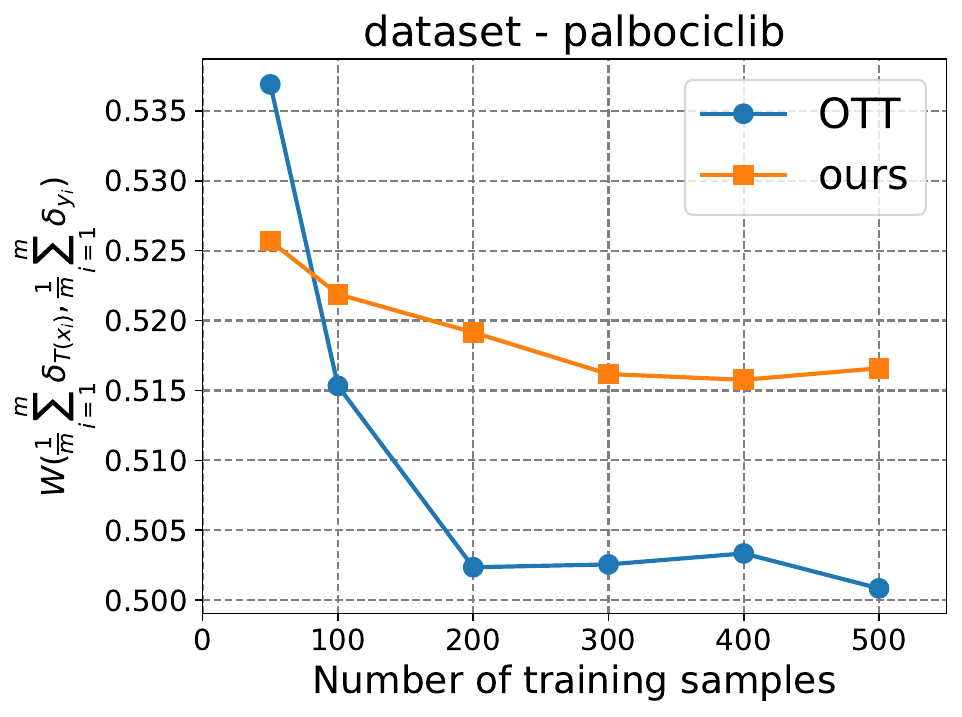}
\caption{Performance of entropic map (using \texttt{OTT}) vs. kernel-based OT estimators computed with the SSN algorithm on 6 drug perturbation datasets. $X$-axis represent the number of training samples and $Y$-axis represents the error induced by OT map $T$ on test samples in terms of OT distance.}\vspace*{-.5em} \label{fig:realdata}
\end{figure*}
\paragraph{Entropic OT vs Kernel OT on Single-cell data.}  Comparing kernel-based OT estimators with plug-in OT estimators on synthetic data has been conducted in~\citet{Vacher-2021-Dimension, Muzellec-2021-Near} and the results show that the kernel-based OT estimators behave better when the number of samples is small. We validate this claim using the real-world 4i datasets from~\citet{Bunne-2021-Learning}, which track unaligned populations of cells \textit{before} and \textit{after} perturbations. Our experiments are conducted on 15 datasets with different drug perturbations. We consider as a baseline approach \citeauthor{Pooladian-2021-Entropic}'s entropic map estimator, as implemented in the \texttt{OTT} package~\citep{Cuturi-2022-OTT}. We use their default implementation, which relies on an adaptive choice for the entropic regularization parameter $\varepsilon$. 

Due to space limits, we only present the results on 6 datasets in Figure~\ref{fig:realdata} and defer the results on other datasets to Appendix~\ref{sec:exp-appendix}. We observe that kernel-based OT estimators (computed using our SSN method) achieve satisfactory performance and behave better when the number of training samples is small; indeed, they are better on 6 datasets, comparable on 5 datasets and worse on 4 datasets. While we do expect that the entropic estimator will eventually scale, and outperform our algorithm as the number of training samples increases, these experiments show that kernel-based OT estimation provides a fairly effective alternative when the number of training samples is small, which is consistent with previous observations on synthetic data~\citep{Vacher-2021-Dimension, Muzellec-2021-Near}. These results therefore validate the sample efficiency of our algorithm for computing kernel-based OT Monge map estimators in small $n$ large $d$ regimes. Note that the performance drop on the \textsf{palbocilib} dataset for large sample sizes agrees with this. To speculate, the larger gap might be because of a low-rank structure within the \textsf{palbocilib} data, which can be better exploited by entropic regularized methods.

%!TEX root = paper.tex
\section{Concluding Remarks}\label{sec:conclu}
We propose a nonsmooth equation model for computing kernel-based OT estimators and show that its special problem structure allows it to be solved in an efficient manner using a SSN method. Specifically, we propose a specialized SSN method that achieves low per-iteration cost by exploiting such structure, and prove a global sublinear rate and a local quadratic rate under standard regularity conditions. Experimental results on synthetic data show that our algorithm is more efficient than the short-step IPM~\citep{Vacher-2021-Dimension}, and the results on real data demonstrate its effectiveness. We hope this progress can motivate further improvements and/or modifications of kernel-based OT approaches.

\section*{Acknowledgements}
This work is supported in part by the Mathematical Data Science program of the Office of Naval Research under grant number N00014-18-1-2764 and by the Vannevar Bush Faculty Fellowship program under grant number N00014-21-1-2941.

%%%%%%%%%%%%%%%%%%%%%%%%%%%%%%%%%%%%%%%%%%%%%%%%%%%%%%%%%%%%

\bibliographystyle{plainnat}
\bibliography{ref}

\clearpage
\appendix
\section{Proof of Proposition~\ref{Prop:equivalence-main}}
We first prove that $\hat{\gamma}$ is an optimal solution of Eq.~\eqref{prob:main} if $\hat{w} = (\hat{\gamma}, \hat{X})$ satisfies $R(\hat{w}) = 0$ for some $\hat{X} \succeq 0$. Indeed, by the definition of $R$ from Eq.~\eqref{def:residue}, we have
\begin{equation}\label{inequality:equivalence-first}
\tfrac{1}{2\lambda_2}Q\hat{\gamma} - \tfrac{1}{2\lambda_2}z - \Phi(\hat{X}) = 0, 
\end{equation}
and 
\begin{equation}\label{inequality:equivalence-second}
\hat{X} - \proj_{\SCal_+^n}(\hat{X} - (\Phi^\star(\hat{\gamma}) + \lambda_1 I)) = 0. 
\end{equation}
By the definition of $\proj_{\SCal_+^n}$, we have
\begin{equation*}
\langle X - \proj_{\SCal_+^n}(\hat{X} - (\Phi^\star(\hat{\gamma}) + \lambda_1 I)), \proj_{\SCal_+^n}(\hat{X} - (\Phi^\star(\hat{\gamma}) + \lambda_1 I)) - \hat{X} + (\Phi^\star(\hat{\gamma}) + \lambda_1 I)\rangle \geq 0 \textnormal{ for all } X \succeq 0. 
\end{equation*}
Plugging Eq.~\eqref{inequality:equivalence-second} into the above inequality yields that 
\begin{equation*}
\langle X - \hat{X}, \Phi^\star(\hat{\gamma}) + \lambda_1 I\rangle \geq 0 \textnormal{ for all } X \succeq 0. 
\end{equation*}
By setting $X = 0$ and $X = 2\hat{X}$, we have $\langle\hat{X}, \Phi^\star(\hat{\gamma}) + \lambda_1 I\rangle \leq 0$ and $\langle\hat{X}, \Phi^\star(\hat{\gamma}) + \lambda_1 I\rangle \geq 0$. Thus, we have 
\begin{equation}\label{inequality:equivalence-third}
\langle\hat{X}, \Phi^\star(\hat{\gamma}) + \lambda_1 I\rangle = 0, \quad \langle X, \Phi^\star(\hat{\gamma}) + \lambda_1 I\rangle \geq 0 \textnormal{ for all } X \succeq 0. 
\end{equation}
Suppose that $\gamma \in \br^n$ satisfies that $\Phi^\star(\gamma) + \lambda_1 I \succeq 0$, we have 
\begin{eqnarray*}
\lefteqn{0 \overset{~\eqref{inequality:equivalence-first}}{=} (\gamma - \hat{\gamma})^\top\left(\tfrac{1}{2\lambda_2}Q\hat{\gamma} - \tfrac{1}{2\lambda_2}z - \Phi(\hat{X})\right)} \\
& = & \left(\tfrac{1}{4\lambda_2}\gamma^\top Q\gamma - \tfrac{1}{2\lambda_2}\gamma^\top z\right) - \left(\tfrac{1}{4\lambda_2}\hat{\gamma}^\top Q\hat{\gamma} - \tfrac{1}{2\lambda_2}\hat{\gamma}^\top z\right) - \tfrac{1}{4\lambda_2}(\gamma - \hat{\gamma})^\top Q(\gamma - \hat{\gamma}) - (\gamma - \hat{\gamma})^\top\Phi(\hat{X}) \\
& \leq & \left(\tfrac{1}{4\lambda_2}\gamma^\top Q\gamma - \tfrac{1}{2\lambda_2}\gamma^\top z\right) - \left(\tfrac{1}{4\lambda_2}\hat{\gamma}^\top Q\hat{\gamma} - \tfrac{1}{2\lambda_2}\hat{\gamma}^\top z\right) - (\gamma - \hat{\gamma})^\top\Phi(\hat{X})
\end{eqnarray*}
Since $\Phi^\star$ is the adjoint of $\Phi$, we have $(\gamma - \hat{\gamma})^\top\Phi(\hat{X}) = \langle \hat{X}, \Phi^\star(\gamma) - \Phi^\star(\hat{\gamma})\rangle$. By combining this equality with $\Phi^\star(\gamma) + \lambda_1 I \succeq 0$ and the first equality in Eq.~\eqref{inequality:equivalence-third}, we have
\begin{equation*}
(\gamma - \hat{\gamma})^\top\Phi(\hat{X}) = \langle \hat{X}, \Phi^\star(\gamma) + \lambda_1 I\rangle - \langle \hat{X}, \Phi^\star(\hat{\gamma}) + \lambda_1 I\rangle \geq 0. 
\end{equation*}
Thus, we have
\begin{equation*}
0 \leq \left(\tfrac{1}{4\lambda_2}\gamma^\top Q\gamma - \tfrac{1}{2\lambda_2}\gamma^\top z + \tfrac{q^2}{4\lambda_2}\right) - \left(\tfrac{1}{4\lambda_2}\hat{\gamma}^\top Q\hat{\gamma} - \tfrac{1}{2\lambda_2}\hat{\gamma}^\top z + \tfrac{q^2}{4\lambda_2}\right). 
\end{equation*}
Combining the above inequality with the second inequality in Eq.~\eqref{inequality:equivalence-third} yields the desired result. 

It suffices to prove that satisfies $R(\hat{w}) = 0$ for some $\hat{X} \succeq 0$ if $\hat{\gamma}$ is an optimal solution of Eq.~\eqref{prob:main}. Indeed, we write that $\sum_{i=1}^n \hat{\gamma}_i\Phi_i\Phi_i^\top + \lambda_1 I \succeq 0$ and 
\begin{equation*}
\begin{array}{r}
\tfrac{1}{4\lambda_2}\hat{\gamma}^\top Q\hat{\gamma} - \tfrac{1}{2\lambda_2}\hat{\gamma}^\top z + \tfrac{q^2}{4\lambda_2} \leq \tfrac{1}{4\lambda_2}\gamma^\top Q\gamma - \tfrac{1}{2\lambda_2}\gamma^\top z + \tfrac{q^2}{4\lambda_2}, \\
\end{array}
\end{equation*}
for all $\gamma \in \br^n$ satisfying that $\sum_{i=1}^n \gamma_i\Phi_i\Phi_i^\top + \lambda_1 I \succeq 0$. Then, the KKT condition guarantees that there exists some $\hat{X} \succeq 0$ satisfying that 
\begin{equation}\label{inequality:equivalence-fourth}
\begin{array}{rcl}
\sum_{i=1}^n \hat{\gamma}_i\Phi_i\Phi_i^\top + \lambda_1 I & \succeq & 0, \\
\tfrac{1}{2\lambda_2}Q\hat{\gamma} - \tfrac{1}{2\lambda_2}z - \Phi(\hat{X}) & = & 0, \\
\langle\hat{X}, \Phi^\star(\hat{\gamma}) + \lambda_1 I\rangle & = & 0. 
\end{array}
\end{equation}
The first and third inequalities guarantee that  
\begin{equation*}
\langle X - \hat{X}, \Phi^\star(\hat{\gamma}) + \lambda_1 I\rangle \geq 0 \textnormal{ for all } X \succeq 0. 
\end{equation*}
By letting $X = \proj_{\SCal_+^n}(\hat{X} - (\Phi^\star(\hat{\gamma}) + \lambda_1 I))$, we have 
\begin{equation}\label{inequality:equivalence-fifth}
\langle \proj_{\SCal_+^n}(\hat{X} - (\Phi^\star(\hat{\gamma}) + \lambda_1 I)) - \hat{X}, \Phi^\star(\hat{\gamma}) + \lambda_1 I\rangle \geq 0. 
\end{equation}
Recall that the definition of $\proj_{\SCal_+^n}$ implies that 
\begin{equation*}
\langle X - \proj_{\SCal_+^n}(\hat{X} - (\Phi^\star(\hat{\gamma}) + \lambda_1 I)), \proj_{\SCal_+^n}(\hat{X} - (\Phi^\star(\hat{\gamma}) + \lambda_1 I)) - \hat{X} + (\Phi^\star(\hat{\gamma}) + \lambda_1 I)\rangle \geq 0 \textnormal{ for all } X \succeq 0.  
\end{equation*}
By letting $X = \hat{X}$, we have 
\begin{equation*}
\|\proj_{\SCal_+^n}(\hat{X} - (\Phi^\star(\hat{\gamma}) + \lambda_1 I)) - \hat{X}\|^2 \leq \langle \hat{X} - \proj_{\SCal_+^n}(\hat{X} - (\Phi^\star(\hat{\gamma}) + \lambda_1 I)), \Phi^\star(\hat{\gamma}) + \lambda_1 I\rangle \overset{~\eqref{inequality:equivalence-fifth}}{\leq} 0. 
\end{equation*}
Combining the above inequality with the second equality in Eq.~\eqref{inequality:equivalence-fourth} yields that 
\begin{equation*}
\tfrac{1}{2\lambda_2}Q\hat{\gamma} - \tfrac{1}{2\lambda_2}z - \Phi(\hat{X}) = 0, \qquad \hat{X} - \proj_{\SCal_+^n}(\hat{X} - (\Phi^\star(\hat{\gamma}) + \lambda_1 I)) = 0. 
\end{equation*}
Combining these inequalities with the definition of $R$ implies $R(\hat{w}) = 0$ and hence the desired result. 

\section{Proof of Proposition~\ref{Prop:strong-semismooth}}
The strong semismoothness of $R$ follows from the derivation given in~\citet{Sun-2002-Semismooth} to establish the semismoothness of projection operators. Indeed, the projection over a positive semidefinite cone is guaranteed to be strongly semismooth~\citep[Corollary~4.15]{Sun-2002-Semismooth}. Thus, we have that $\proj_{\SCal_+^n}(\cdot)$ is strongly semismooth. Since the strong semismoothness is closed under scalar multiplication, summation and composition, the residual map $R$ is strongly semismooth. 

\section{Proof of Lemma~\ref{Lemma:key-structure}}\label{app:key-structure}
As stated in Lemma~\ref{Lemma:key-structure}, we compute $Z_k = X_k - (\Phi^\star(\gamma_k) + \lambda_1 I)$ and the spectral decomposition of $Z_k$ (cf. Eq.~\eqref{def:SD}) to obtain $P_k$, $\Sigma_k$ and the sets of the indices of positive and nonpositive eigenvalues $\alpha_k$ and $\bar{\alpha}_k$. We then compute $\Omega_k$ using $\Sigma_k$, $\alpha_k$ and $\bar{\alpha}_k$ and finally obtain that $\tilde{P}_k = P_k \otimes P_k$ and $\Gamma_k = \diag(\textnormal{vec}(\Omega_k))$. Thus, we can write the matrix form of $\JCal_k + \mu_k I$ as
\begin{equation*}
J_k + \mu_k I = \begin{pmatrix} \tfrac{1}{2\lambda_2}Q + \mu_k I & -A \\ \tilde{P}_k\Gamma_k\tilde{P}_k^\top A^\top & \tilde{P}_k((\mu_k+1)I - \Gamma_k)\tilde{P}_k^\top \end{pmatrix}. 
\end{equation*}
For simplicity, we let $W_k = \tilde{P}_k\Gamma_k\tilde{P}_k^\top$ and $D_k = \tilde{P}_k((\mu_k+1)I - \Gamma_k)\tilde{P}_k^\top$. Then, the Schur complement trick implies that 
\begin{eqnarray*}
\lefteqn{(J_k + \mu_k I)^{-1} = \begin{pmatrix} \tfrac{1}{2\lambda_2}Q + \mu_k I & -A \\ W_k A^\top & D_k \end{pmatrix}^{-1}} \\
& = & \begin{pmatrix} I & 0 \\ -D_k^{-1}W_kA^\top & I \end{pmatrix}\begin{pmatrix} (\tfrac{1}{2\lambda_2}Q + \mu_k I + AD_k^{-1}W_kA^\top)^{-1} & 0 \\ 0 & D_k^{-1} \end{pmatrix}\begin{pmatrix} I & AD_k^{-1} \\ 0 & I \end{pmatrix}. 
\end{eqnarray*}
Define $T_k = \tilde{P}_k L_k\tilde{P}_k^\top$ where $L_k$ is a diagonal matrix with $(L_k)_{ii} = \frac{(\Gamma_k){ii}}{\mu_k + 1 - (\Gamma_k)_{ii}}$ and $(\Gamma_k)_{ii} \in (0, 1]$ is the $i^\textnormal{th}$ diagonal entry of $\Gamma_k$. By the definition of $W_k$ and $D_k$, we have $D_k^{-1} = \tfrac{1}{\mu_k + 1}(I + T_k)$ and $D_k^{-1}W = T_k$. Using these two identities, we can further obtain that 
\begin{equation*}
(J_k + \mu_k I)^{-1} = \begin{pmatrix} I & 0 \\ -T_k A^\top & I \end{pmatrix}\begin{pmatrix} (\tfrac{1}{2\lambda_2}Q + \mu_k I + AT_kA^\top)^{-1} & 0 \\ 0 & \tfrac{1}{\mu_k + 1}(I + T_k) \end{pmatrix}\begin{pmatrix} I & \tfrac{1}{\mu_k + 1}(A + AT_k) \\ 0 & I \end{pmatrix}. 
\end{equation*}
This completes the proof. 

\section{Proof of Theorem~\ref{Thm:convergence-global}}
We can see from the scheme of Algorithm~\ref{alg:main} that 
\begin{equation*}
\|R(w_k)\| \leq \|R(v_k)\| \quad \textnormal{for all } k \geq 0, 
\end{equation*}
where the iterates $\{v_k\}_{k \geq 0}$ are generated by applying the extragradient (EG) method for solving the min-max optimization problem in Eq.~\eqref{prob:minimax}. We also have that~\citet[Theorem~3]{Cai-2022-Finite} guarantees that $\|R(v_k)\| = O(1/\sqrt{k})$. Putting these pieces together yields that 
\begin{equation*}
\|R(w_k)\| = O(1/\sqrt{k}). 
\end{equation*}
This completes the proof. 
\section{Proof of Theorem~\ref{Thm:convergence-local}}\label{app:convergence-local}
We analyze the convergence property for one-step SSN step as follows, 
\begin{equation*}
w_{k+1} = w_k + \Delta w_k, 
\end{equation*}
where $\mu_k = \theta_k\|R(w_k)\|$ and 
\begin{equation}\label{inequality:local-first}
\|(\JCal_k + \mu_k\ICal)[\Delta w_k] + R(w_k)\| \leq \tau\min\{1, \kappa\|R(w_k)\|\|\Delta w_k\|\}. 
\end{equation}
Since $R$ is strongly smooth (cf. Proposition~\ref{Prop:strong-semismooth}), we have
\begin{equation*}
\tfrac{\|R(w + \Delta w) - R(w) - \JCal[\Delta w]\|}{\|\Delta w\|^2} \leq C, \quad \textnormal{as } \Delta w \rightarrow 0. 
\end{equation*}
Since $w_0$ is sufficiently close to $w^\star$ with $R(w^\star) = 0$ and the global convergence guarantee holds (cf. Theorem~\ref{Thm:convergence-global}), we have
\begin{equation*}
\|R(w_k + \Delta w_k) - R(w_k) - \JCal_k[\Delta w_k]\| \leq 2C\|\Delta w_k\|^2. 
\end{equation*}
which implies that 
\begin{equation}\label{inequality:local-second}
\|R(w_{k+1})\| = \|R(w_k + \Delta w_k)\| \leq \|R(w_k) + \JCal_k[\Delta w_k]\| + 2C\|\Delta w_k\|^2. 
\end{equation}
Plugging Eq.~\eqref{inequality:local-first} into Eq.~\eqref{inequality:local-second} yields that 
\begin{eqnarray}\label{inequality:local-third}
\|R(w_{k+1})\| & \leq & 2C\|\Delta w_k\|^2 + \mu_k\|\Delta w_k\| + \tau\kappa\|R(w_k)\|\|\Delta w_k\| \\ 
& \leq & 2C\|\Delta w_k\|^2 + (\theta_k + \tau\kappa)\|R(w_k)\|\|\Delta w_k\|. \nonumber
\end{eqnarray}
Since $w_0$ is sufficiently close to $w^\star$ with $R(w^\star) = 0$ and every element of $\partial R(w^\star)$ is invertible, we have that there exists some $\delta > 0$ such that 
\begin{equation*}
\|(\JCal_k + \mu_k\ICal)[\Delta w_k]\| \geq \delta\|\Delta w_k\|. 
\end{equation*}
The above equation together with Eq.~\eqref{inequality:local-first} yields that 
\begin{equation}\label{inequality:local-fourth}
\|\Delta w_k\| \leq \tfrac{1}{\delta}\|(\JCal_k + \mu_k\ICal)[\Delta w_k]\| \leq \tfrac{1}{\delta}\left(1 + \tau\kappa\|\Delta w_k\|\right)\|R(w_k)\|. 
\end{equation}
Plugging Eq.~\eqref{inequality:local-fourth} into Eq.~\eqref{inequality:local-third} yields that 
\begin{equation*}
\|R(w_{k+1})\| \leq \|R(w_k)\|^2\left(\tfrac{2C}{\delta^2}\left(1 + \tau\kappa\|\Delta w_k\|\right)^2 + \tfrac{\theta_k + \tau\kappa}{\delta}\left(1 + \tau\kappa\|\Delta w_k\|\right)\right)
\end{equation*}
Note that $\|\Delta w_k\| \rightarrow 0$ and $\theta_k$ is bounded. Thus, we have $\|R(w_{k+1})\| = O(\|R(w_k)\|^2)$. 
\begin{figure*}[!t]
\centering
\includegraphics[width=0.3\textwidth]{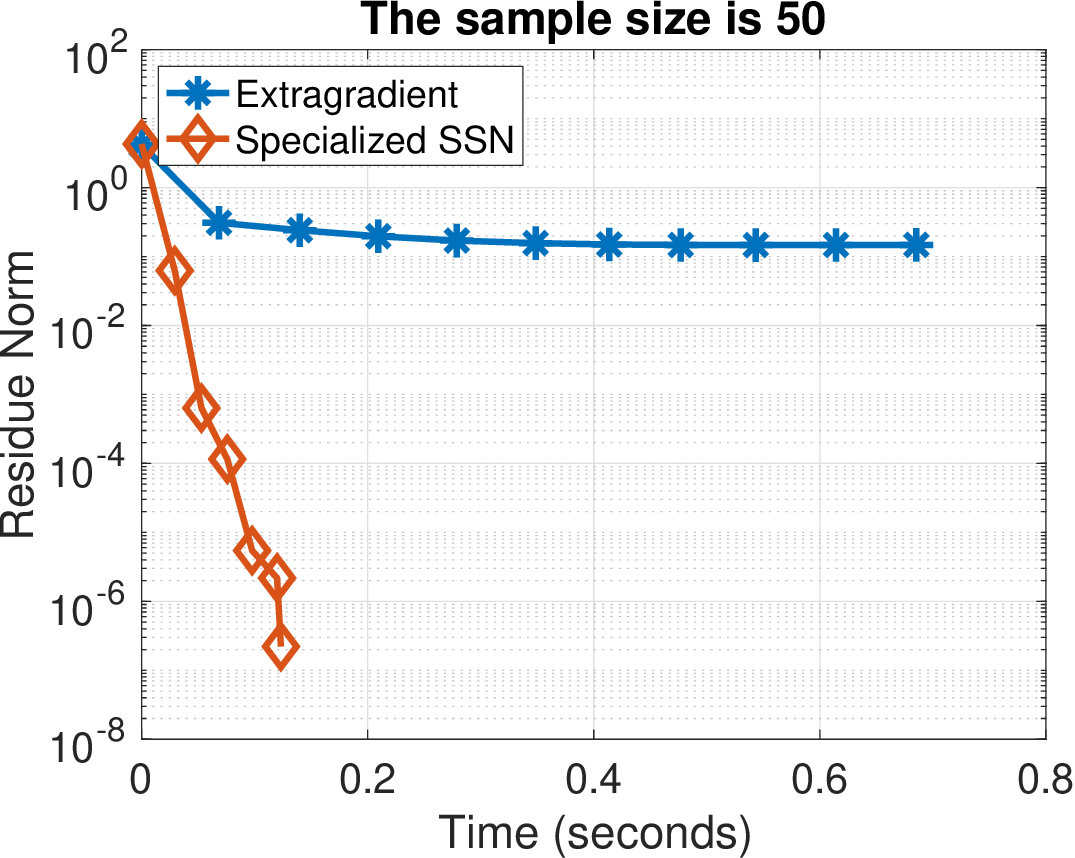}
\includegraphics[width=0.3\textwidth]{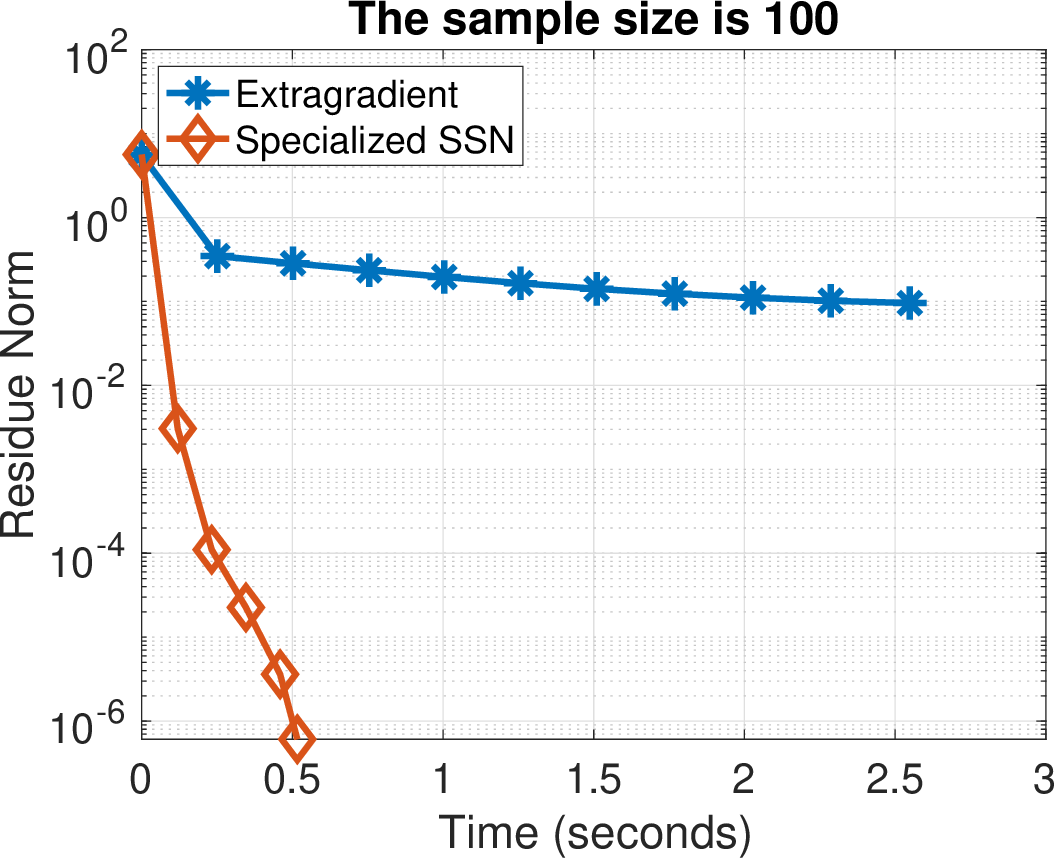}
\includegraphics[width=0.3\textwidth]{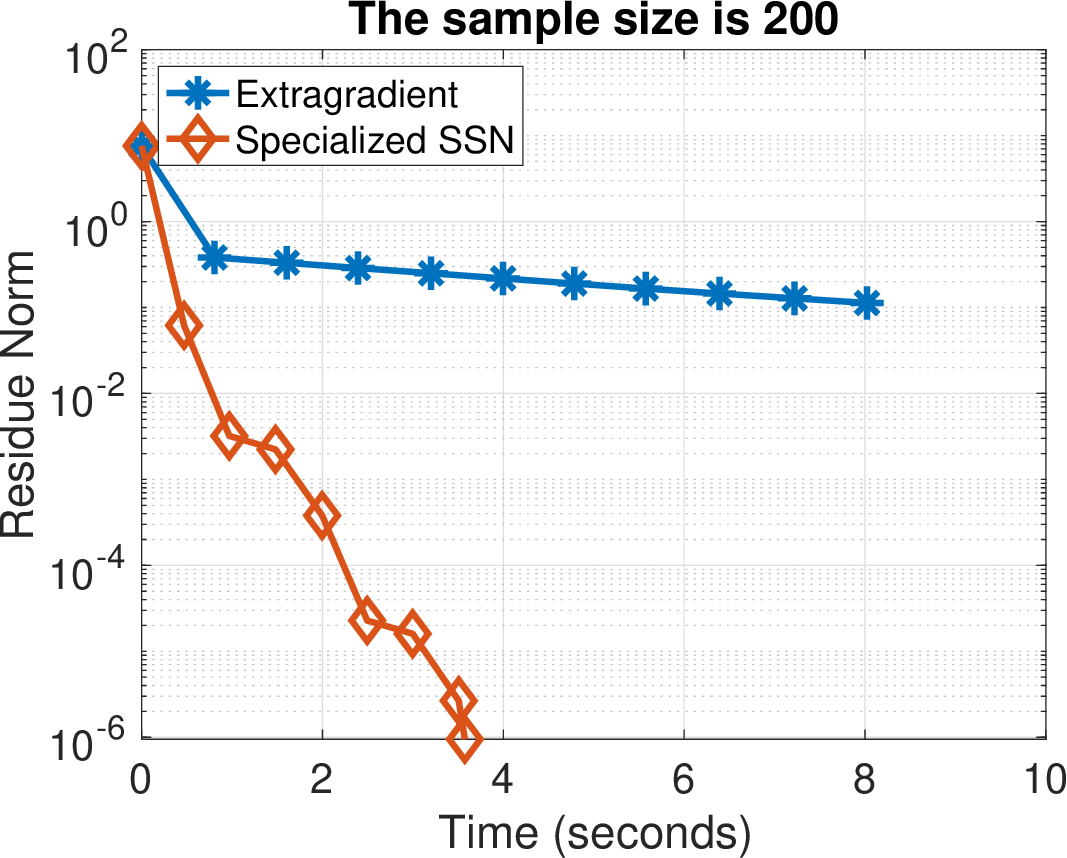} \\
\includegraphics[width=0.3\textwidth]{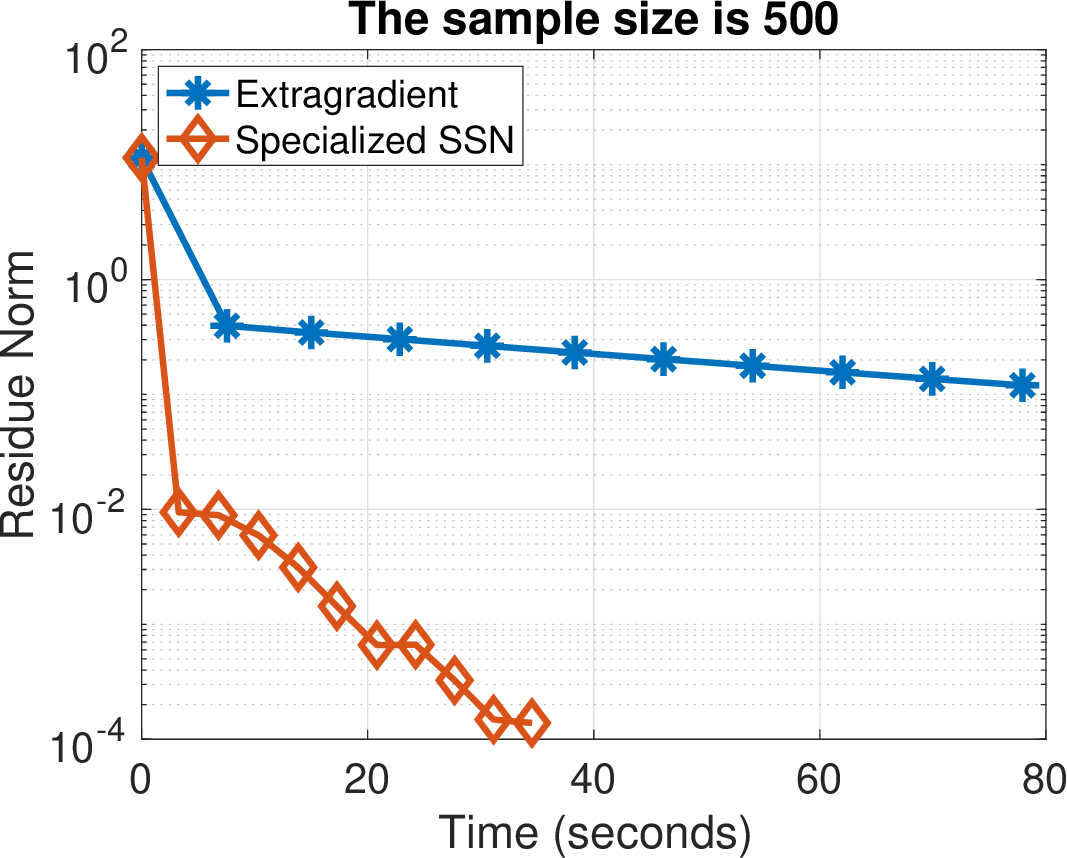}
\includegraphics[width=0.3\textwidth]{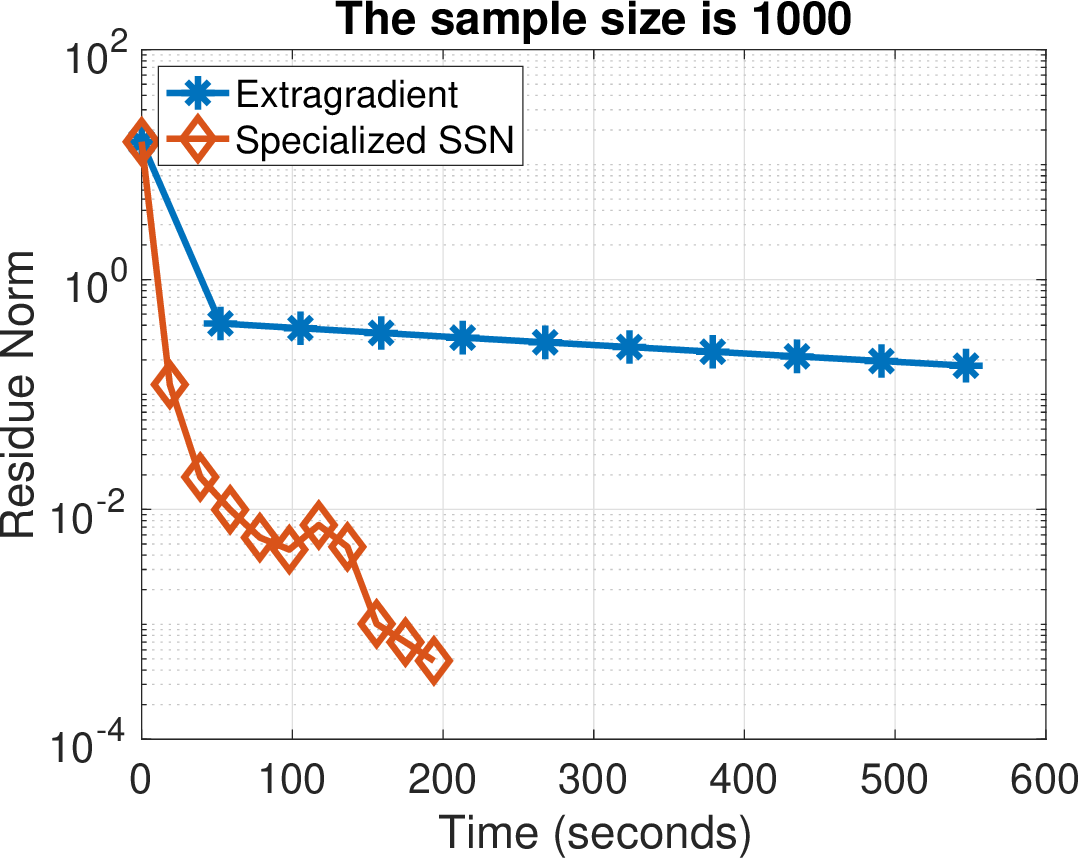}
\includegraphics[width=0.3\textwidth]{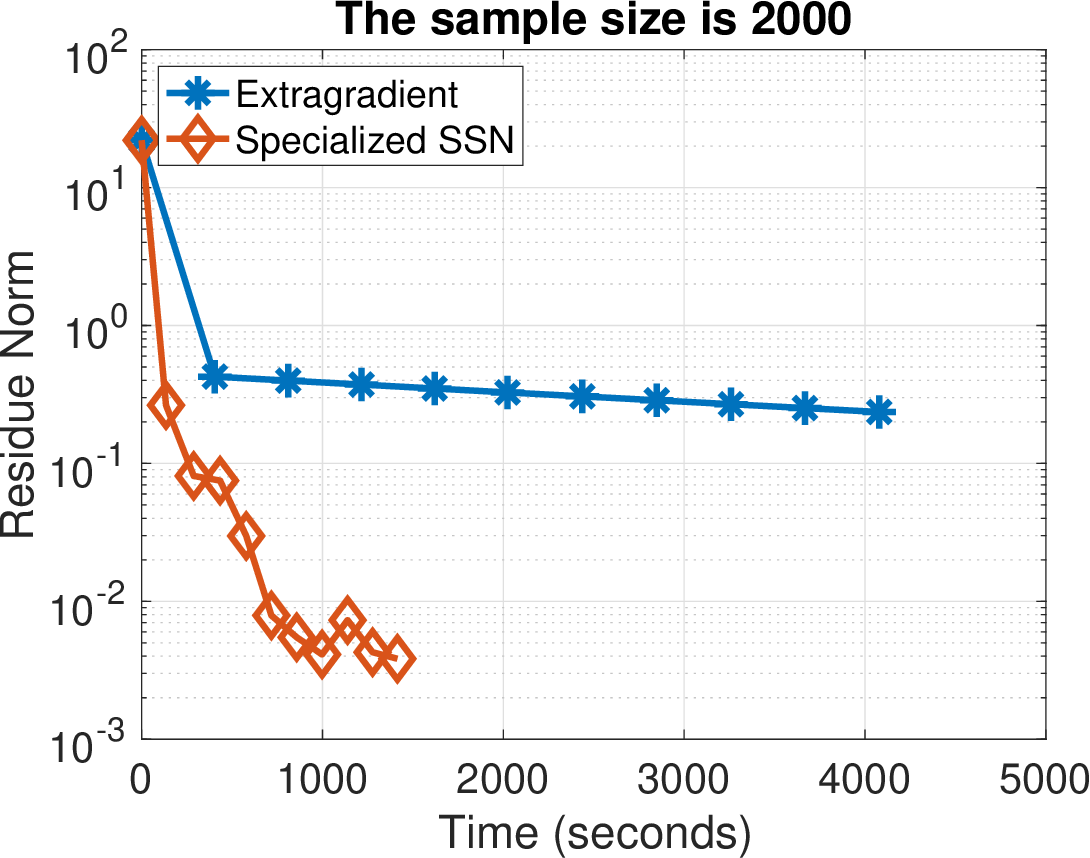}
\vspace*{-.5em}\caption{\footnotesize{Performance of pure EG and our algorithm for solving kernel-based OT problems with the varying sample size $n \in \{50, 100, 200, 500, 1000, 2000\}$. The numerical results are presented as residue norm v.s. time (seconds).}}
\label{fig:syntheticdata-appendix}
\end{figure*}
From the above arguments, we see that the quadratic convergence rate can be achieved if Algorithm~\ref{alg:main} performs the SSN step when the initial iterate $x_0$ is sufficiently close to $w^\star$ with $R(w^\star) = 0$. This implies that the safeguarding steps will never affect in local sense where Algorithm~\ref{alg:main} generates $\{w_k\}_{k \geq 0}$ by performing the SSN steps only. So Algorithm~\ref{alg:main} achieves the local quadratic convergence.  This completes the proof. 

\section{Additional Experiments}\label{sec:exp-appendix}
We compare our method with the pure extragradient (EG) method and summarize the numerical results in Figure~\ref{fig:syntheticdata-appendix}. In particular, we find that our method consistently outperforms the pure EG method and can output a high-accurate solution in terms of the residue norm. The experimental setup is the same as that used in the main context. Indeed, we fix the dimension $d = 10$ and the bandwidth $\sigma^2 = 0.005$, and vary the sample size $n \in \{50, 100, 200, 500, 1000, 2000\}$. For the EG method, we tune the stepsize and set it as $0.01$.  

We also describe our setup for the experiment on the real-world 4i datasets from~\citet{Bunne-2021-Learning}. Indeed, we draw the unperturbed/perturbed samples for training from 15 cell datasets as follows, 
\begin{equation*}
x_1, \ldots, x_{n_\textnormal{sample}} \sim \mu_{\text{unperturb}}, \quad y_1, \ldots, y_{n_\textnormal{sample}} \sim \nu_{\text{perturb}}^{k} \textnormal{ for } 1 \leq k \leq 15. 
\end{equation*}
where $x_i, y_i \in \br^{48}$ and $\mu_{\text{unperturb}}, \nu_{\text{perturb}}^{k}$ represent the unperturbed cells and $k^\textnormal{th}$ perturbed cells. For our algorithm, we generate $256$ filling points and compare our method with the default implementation in \texttt{OTT} package~\citep{Cuturi-2022-OTT}. Here, the value of entropic parameter is automatically selected by \texttt{OTT} package. 

\begin{figure*}[!t]
\centering
\includegraphics[width=.3\textwidth]{figs/plot_X-perturbed-drug-cisplatin_olaparib.pdf}
\includegraphics[width=.3\textwidth]{figs/plot_X-perturbed-drug-decitabine.pdf}
\includegraphics[width=.3\textwidth]{figs/plot_X-perturbed-drug-trametinib_panobinostat.pdf}
\includegraphics[width=.3\textwidth]{figs/plot_X-perturbed-drug-cisplatin.pdf}
\includegraphics[width=.3\textwidth]{figs/plot_X-perturbed-drug-sorafenib.pdf}
\includegraphics[width=.3\textwidth]{figs/plot_X-perturbed-drug-palbociclib.pdf}
\includegraphics[width=.3\textwidth]{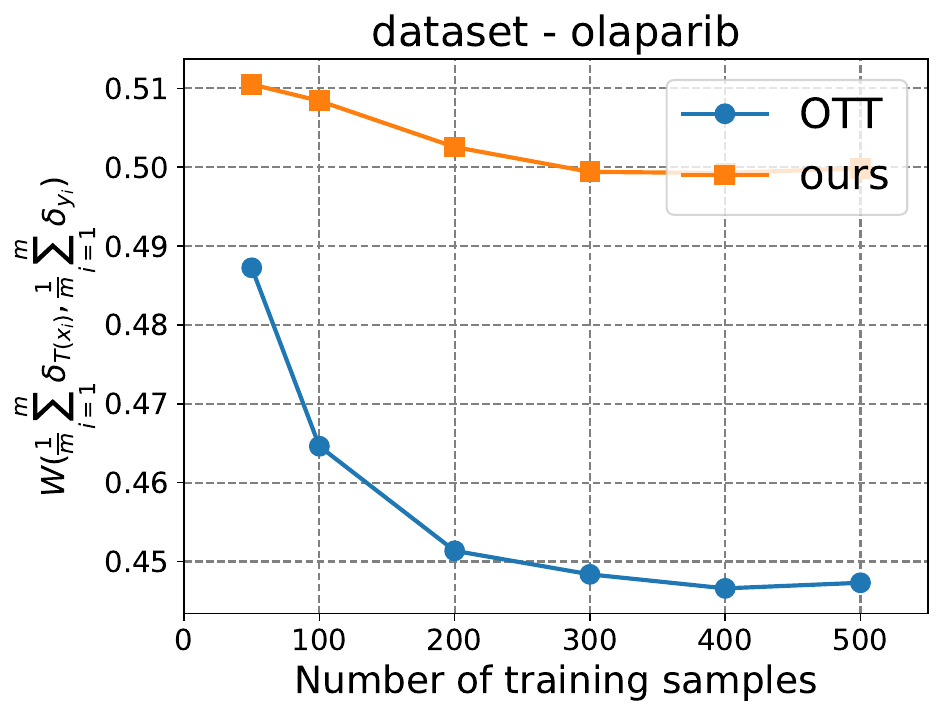}
\includegraphics[width=.3\textwidth]{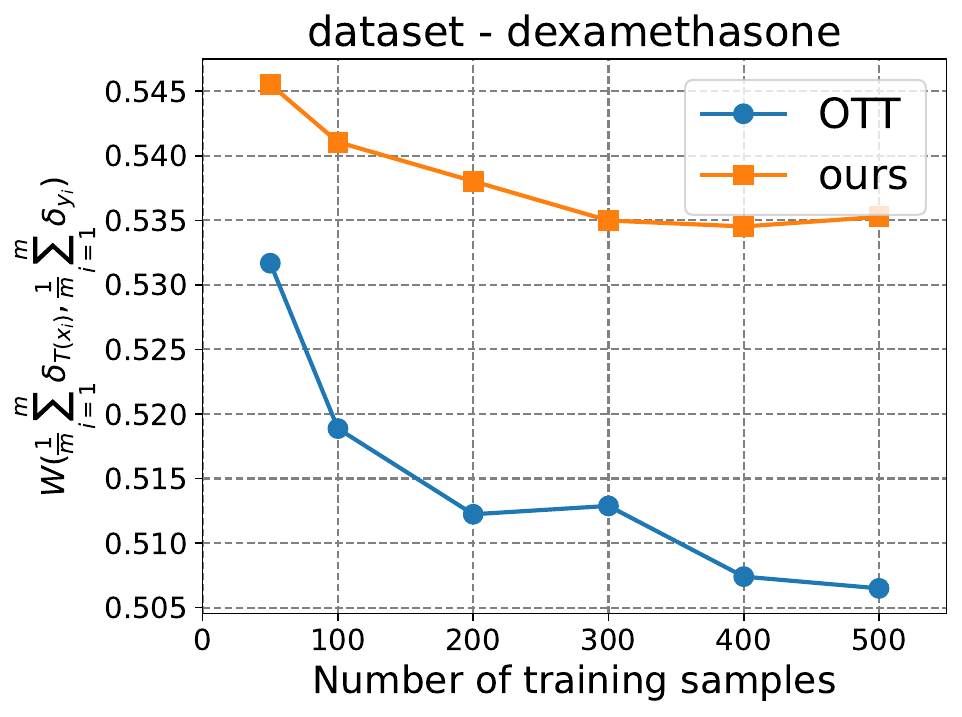}
\includegraphics[width=.3\textwidth]{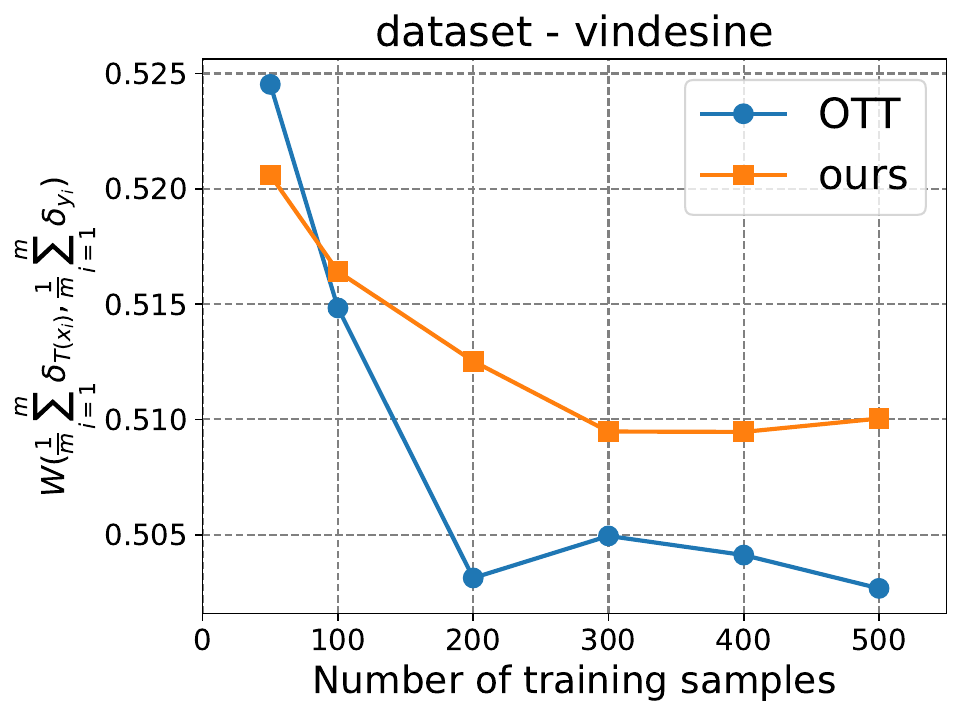}
\includegraphics[width=.3\textwidth]{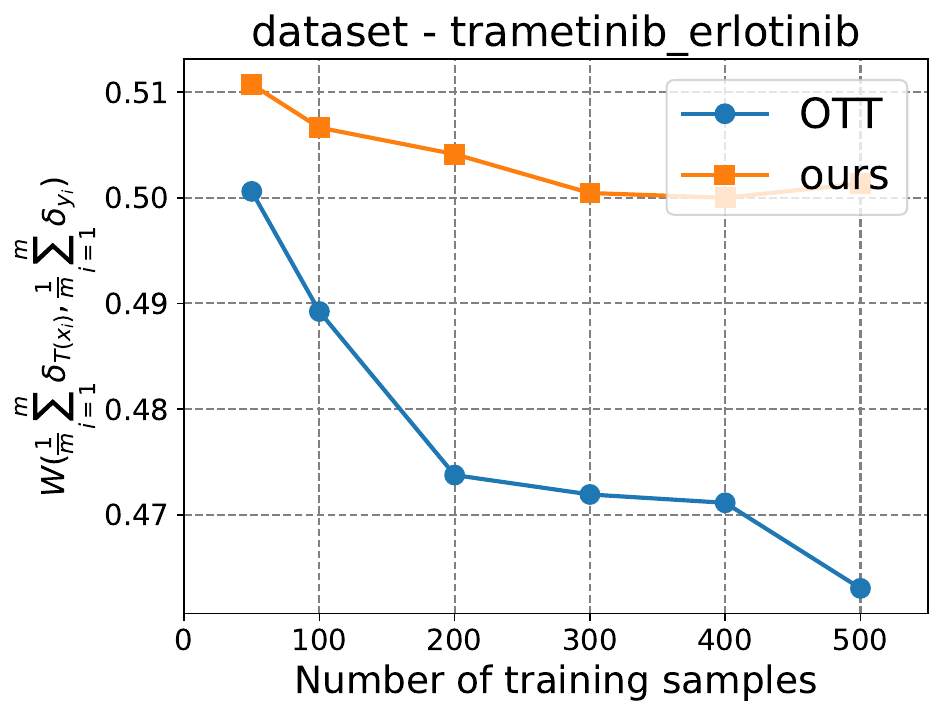}
\includegraphics[width=.3\textwidth]{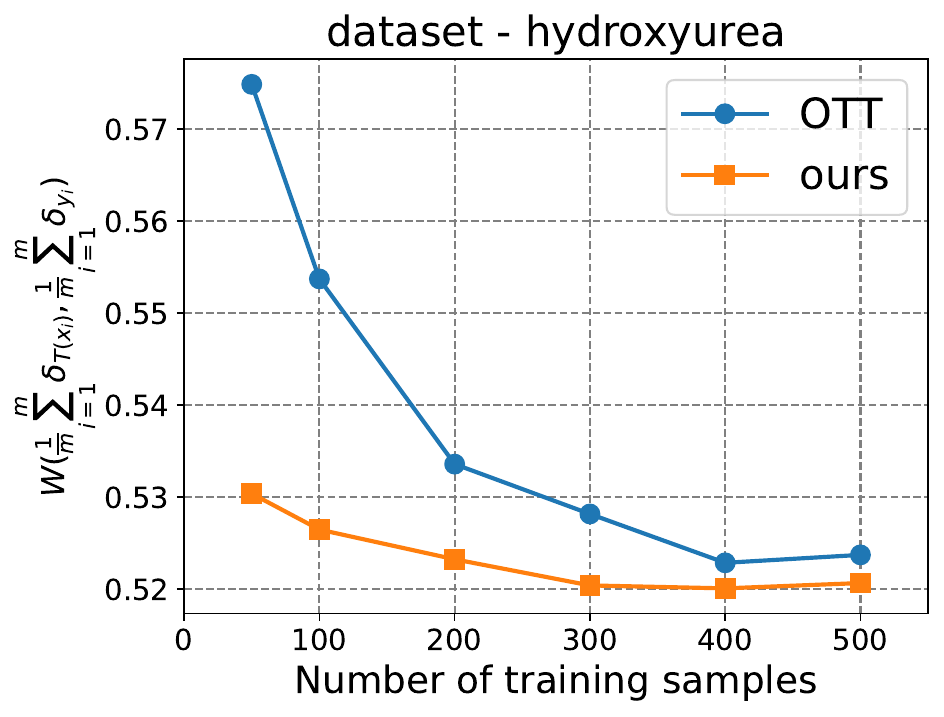}
\includegraphics[width=.3\textwidth]{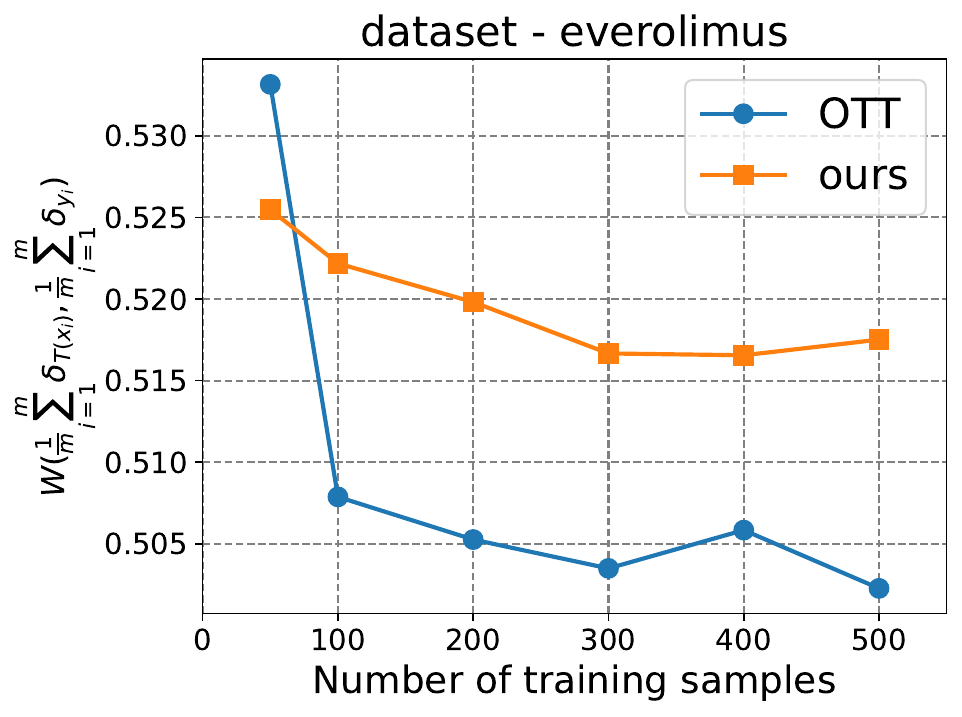}
\includegraphics[width=.3\textwidth]{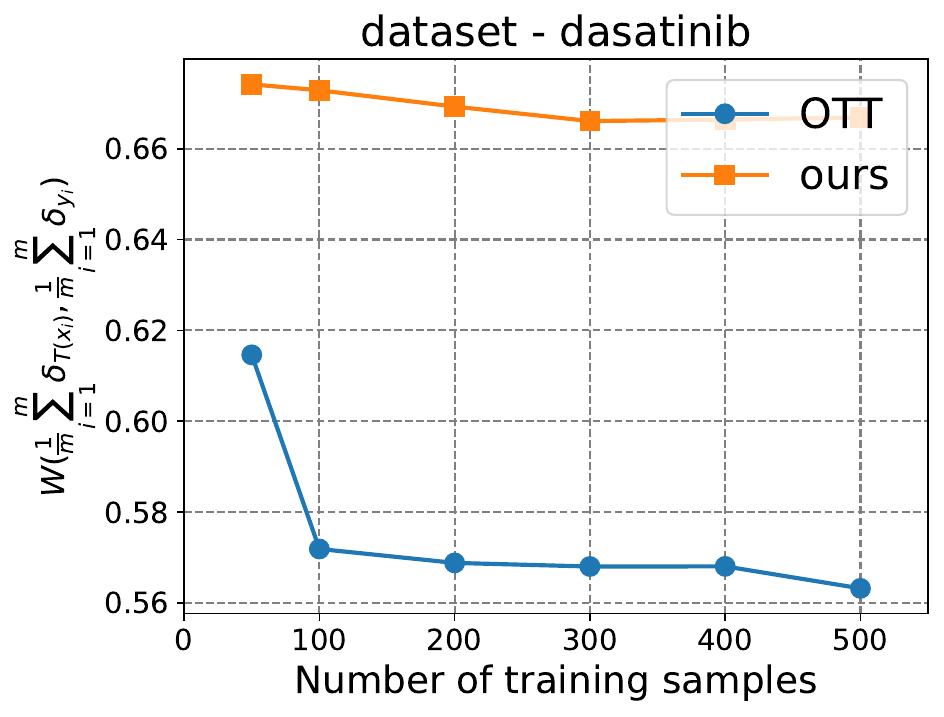}
\includegraphics[width=.3\textwidth]{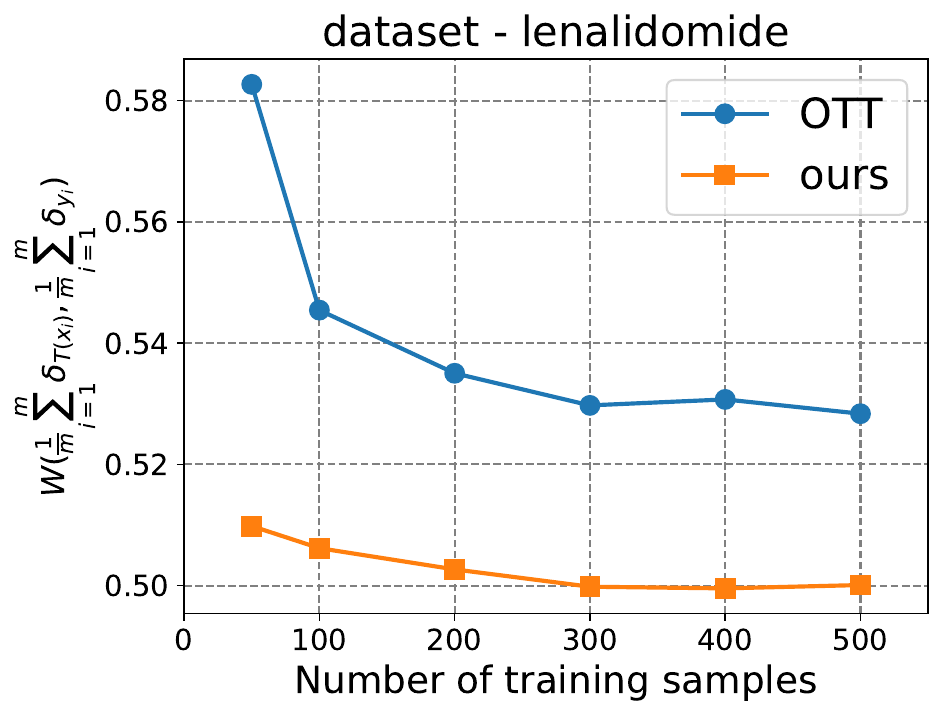}
\includegraphics[width=.3\textwidth]{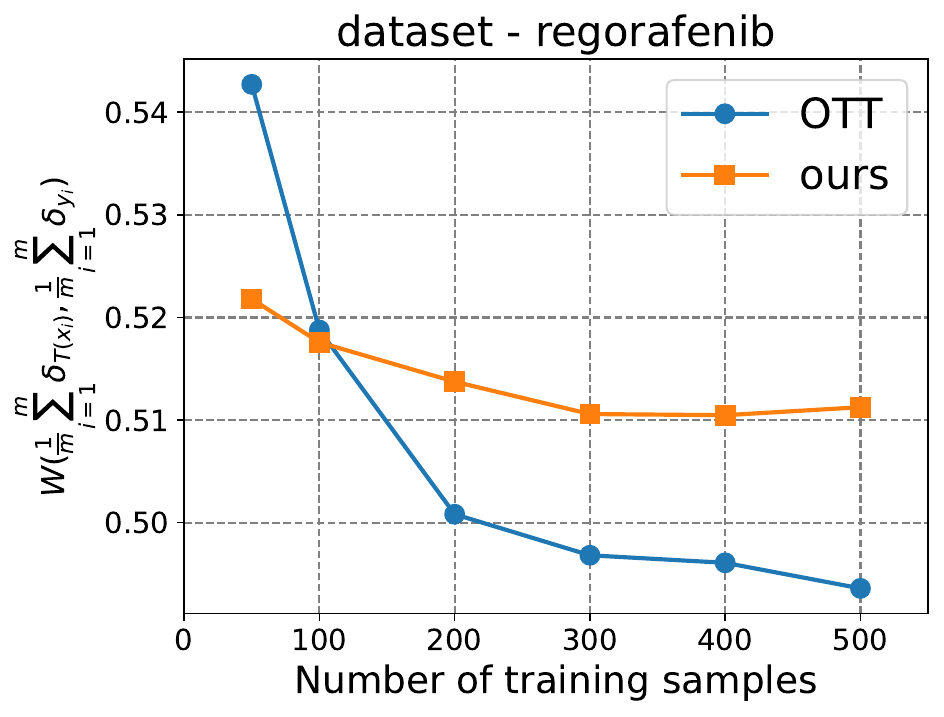}
\caption{\footnotesize{Performance of \texttt{OTT} and kernel-based OT estimators computed by our algorithm on all of 15 drug perturbation datasets. $X$-axis represent the number of training samples and $Y$-axis represents the error induced by OT map $T$ on test samples in terms of OT distance.}} 
\label{fig:realdata-appendix}
\end{figure*}

Both our algorithm and \texttt{OTT} capture the OT map $T$ from training samples. Then, we fix the number of test samples as $m=200$ and use the OT distance to measure the differences between $\frac{1}{m}\sum_{j=1}^{m}\delta_{T(\hat{x}_j)}$ and $\frac{1}{m}\sum_{j=1}^{m}\delta_{\hat{y}_j}$, where $\hat{x}_1, \dots, \hat{x}_m \sim \mu_{\text{unperturb}}$ and $\hat{y}_1, \dots, \hat{y}_m \sim \nu_{\text{perturb}}^k$ are unperturbed/perturbed samples for testing. Figure~\ref{fig:realdata-appendix} reports the results on 15 single-cell datasets.

\end{document}